\documentclass{article}

\usepackage{PRIMEarxiv}

\usepackage[utf8]{inputenc} % allow utf-8 input
\usepackage[T1]{fontenc}    % use 8-bit T1 fonts
\usepackage{url}            % simple URL typesetting
\usepackage{booktabs}       % professional-quality tables
\usepackage{amsfonts}       % blackboard math symbols
\usepackage{nicefrac}       % compact symbols for 1/2, etc.
\usepackage{microtype}      % microtypography
\usepackage{lipsum}
\usepackage{fancyhdr}       % header
\usepackage{graphicx}       % graphics
\graphicspath{{media/}}     % organize your images and other figures under media/ folder

\usepackage{adjustbox}

\usepackage[title]{appendix}

\usepackage{array}

\usepackage{algpseudocode}

\usepackage{algorithm}

\usepackage{url}
\usepackage{caption}
\usepackage{subcaption}

\usepackage{lineno}
\usepackage{amsmath}

\usepackage{hyperref}       % hyperlinks

\DeclareMathOperator*{\argmin}{argmin}
\DeclareMathOperator*{\argmax}{argmax}
\newcommand{\X}{{\mathbf{X}}}
\newcommand{\Y}{{\mathbf{Y}}}
\newcommand{\Dx}{{\mathbf{D_x}}}
\newcommand{\Dy}{{\mathbf{D_y}}}

\newcommand{\Rea}{{\mathbb{R}}}

\usepackage{amsthm}
\usepackage{bm}
\usepackage{bbm}
\theoremstyle{thmstyleone}%
\newtheorem{theorem}{Proposition}%  meant for continuous numbers
\theoremstyle{thmstyletwo}%

\theoremstyle{thmstylethree}%
\title{Minimal Learning Machine for Multi-Label Learning}

\author{
  Joonas Hämäläinen \\
  University of Jyvaskyla \\
  Faculty of Information Technology\\
  Finland\\
  \texttt{joonas.k.hamalainen@jyu.fi} \\
  \And
  Antoine Hubermont \\
  University of Namur, Namur \\
  Telespazio Belgium, Bastogne\\
  Belgium\\
  \texttt{antoine.hubermont@unamur.be} \\
   \And
  Amauri Souza \\
  Federal Institute of Education, \\Science and Technology of Cear\'{a}---IFCE \\
  Department of Computer Science\\
  Brazil\\
  \texttt{amauriholanda@ifce.edu.br} \\
  \And
  C\'{e}sar L. C. Mattos \\
  Federal University of Cear\'{a}---UFC \\
  Department of Computer Science\\
  Brazil\\
  \texttt{cesarlincoln@dc.ufc.br} \\
  \And
  Jo\~{a}o P. P. Gomes \\
  Federal University of Cear\'{a}---UFC \\
  Department of Computer Science\\
  Brazil\\
  \texttt{jpaulo@dc.ufc.br} \\
  \And
  Tommi Kärkkäinen \\
  University of Jyvaskyla \\
  Faculty of Information Technology\\
  Finland\\
  \texttt{tommi.karkkainen@jyu.fi} \\
}

\begin{document}
\maketitle

\begin{abstract}
Distance-based supervised method, the minimal learning machine, constructs a predictive model from data by learning a mapping between input and output distance matrices. In this paper, we propose new methods and evaluate how their core component, the distance mapping, can be adapted to multi-label learning. The proposed approach is based on combining the distance mapping with an inverse distance weighting. Although the proposal is one of the simplest methods in the multi-label learning literature, it achieves state-of-the-art performance for small to moderate-sized multi-label learning problems. In addition to its simplicity, the proposed method is fully deterministic: Its hyper-parameter can be selected via ranking loss-based statistic which has a closed form, thus avoiding conventional cross-validation-based hyper-parameter tuning. In addition, due to its simple linear distance mapping-based construction, we demonstrate that the proposed method can assess the uncertainty of the predictions for multi-label classification, which is a valuable capability for data-centric machine learning pipelines.
\end{abstract}




\section{Introduction}\label{sec1}

%\textcolor{red}{NOTE: Latest working name for the proposed method is \mname. Please use "\textbackslash mname" command for refering to the method.} 

% What is multi-label classification? Cite applications and highlight the importance of the task!!
Multi-label classification (MLC, \cite{tsoumakas2007multi, zhang2013review,bogatinovski2022comprehensive}) refers to a supervised machine learning task in which an instance can be associated with more than one class. %Unlike 
This is different from 
the more common %well-known 
machine learning task, the single-label classification (SLC), where each instance is associated with one %relevant label 
class\textemdash and represented with one active label\textemdash 
only. This simplification means that all the other possible class labels are irrelevant, which can be a naive assumption since many real-world problems are inherently multi-label. 

%Indeed, there exists a multitude of relevant application areas of multi-label classification. These include (but are not limited to) the categorization of texts and documents \cite[e.g.,][]{liu2015multi,jiang2017multi,almeida2018emotion,parwez2019multi,du2019ml,zhang2020multi,chen2022multi}, medical imaging \cite[e.g.,][]{chen2019deep,chen2020deep,pan2020multi,bi2020multi}, health and bioinformatics \cite[e.g.,][]{pliakos2019predicting,chougrad2020multi,zhou2021application,elujide2021application,cai2020multi}, remote sensing \cite[e.g.,][]{kumar2021multilabel}, cybersecurity \cite[e.g.,][]{wang2020locational}, brain research \cite[e.g.,][]{sun2020multi}, power load monitoring \cite[e.g.,][]{li2021non}, and
Indeed, there exists a multitude of relevant application areas of multi-label classification. These include (but are not limited to) the categorization of texts and documents \cite{liu2015multi,jiang2017multi,almeida2018emotion,parwez2019multi,du2019ml,zhang2020multi,chen2022multi}, medical imaging \cite{chen2019deep,chen2020deep,pan2020multi,bi2020multi}, health and bioinformatics \cite{pliakos2019predicting,chougrad2020multi,zhou2021application,elujide2021application,cai2020multi}, remote sensing \cite{kumar2021multilabel}, cybersecurity \cite{wang2020locational}, brain research \cite{sun2020multi}, power load monitoring \cite{li2021non}, and
applications in computational chemistry \cite{maser2021multilabel,saini2022predicting}.
%Orig. list of application from Joonas, included in above
%For instance, text categorization \cite{jiang2017multi}, sentiment analysis \cite{almeida2018emotion}, image annotation \cite{nasierding2009clustering}, and medical diagnosis \cite{chougrad2020multi}. 

Formally, in the MLC, each instance is associated with a bipartition of the possible labels into a set of relevant and irrelevant labels. Typically, an MLC method predicts this bipartition by scoring and/or ranking the labels with thresholding. Therefore, the MLC is closely related to the multi-label ranking. Together, these two tasks are below the umbrella concept of multi-label learning \cite{madjarov2012extensive}.

% Here, we want to have a high-level idea of how the existing approaches tackle MLC.
Typically, conventional supervised methods are not directly applicable to MLC problems. The two main approaches to handling these problems are \cite{tsoumakas2007multi} $i)$ algorithm adaptation and $ii)$ problem transformation. In the first one, an SLC approach is modified %for the MLC, referring to that the SLC approach is adjusted so 
in such a way 
that the correlating labels 
in the MLC are %somehow 
managed in a specific way %handled 
in training. In the second approach, the label space is transformed to be suitable for any SLC method. Such a step may consider a second layer of hyper-parameters, in addition to the core SLC method's hyper-parameters.

%Fundamentally we are interested in whether a MLC approach is better than BR based approach in terms of accuracy or efficiency. In other words, is it beneficial to model the label dependencies.

% Mention problems with the existing approaches that the proposal could solve/address...or what could be improved in the existing approaches? The predictive performance is low?! time complexity? not scalable to high number of outputs? Have less parameters? Easier to train? the learning problem is convex?? 

% In this paper, we introduce ... 

% Describe the empirical results briefly...

% Claims and contributions
In this paper, we evaluate distance regression-based methods for the MLC problems. Our main claim of the paper is that linear distance mapping is a beneficial construction for multi-label learning. The main proposal is to construct a multi-label classifier by integrating the distance regression step from the minimal learning machine (MLM) \cite{de2015minimal} and the inverse distance weighting (IDW) method \cite{shepard1968two}. This combination, referred to as multi-label minimal learning machine (ML-MLM), constructs a simple MLC method that is fully deterministic and whose hyper-parameter can be selected via ranking loss-based statistic which has a closed form. In addition, we demonstrate that ML-MLM can intuitively assess the uncertainty of prediction, which we see as a valuable property for data-centric machine learning pipelines.

% Evidence
In order to support our main claim and demonstrate the viability of the proposed methods, we present theoretical and extensive experimental results. Firstly, we show theoretical results to support and motivate our method's integration of the linear distance mapping and IDW scheme. Secondly, we show an extensive experimental study (ten benchmark datasets, eight core MLC performance metrics) regarding four distance regression-based variants and compared our results with the state-of-the-art. We focus on comparing our methods to a random forest based method (random forest of predictive clustering trees) \cite{Kocev2007rfpct}, which has been one of the best performing methods in extensive experimental comparisons \cite{madjarov2012extensive,bogatinovski2022comprehensive}. The experimental results indicate that ML-MLM and other distance regression-based methods are competitive with the state-of-the-art for small to moderate-sized MLC problems, besides being one of the simplest methods in the MLC literature. In addition, according to the experimental results, the distance regression-based methods clearly outperform multi-label kNN (ML-kNN) \cite{zhang2007ml}.

% Most closely related contributions
Bemporad \cite{Bemporad2020idw} integrated IDW and radial basis functions (RBF) to probe expensive objective function evaluations in the field of global optimization. This integration of IDW and RBF has some methodological similarity with ML-MLM since RBF and MLM are closely related \cite{hamalainen2020minimal}. Another closely related contribution to ours is from Zhang and Zhou \cite{zhang2007ml}. This paper proposed ML-kNN, which is similar to our proposal, because ML-kNN also uses feature space distances for multi-label learning. In the MLC taxonomy, ML-kNN belongs to the same category with our method; however, the Bemporad's contribution is closer to ours. Moreover, this paper was inspired by the promising preliminary results of MLM in MLC \cite{hamalainen2021instance}.

% How new methods differ from previous ones and highligihting the contributions of this paper
The method closest to the proposed main method
is the NN-MLM approach \cite{Mesquita2017}. However, the proposed method clearly differs from NN-MLM. First, the proposed method is able to provide ranking of the labels, thus extending MLM for multi-label learning. Secondly, the hyper-parameter selection is designed with the ranking loss based statistic which was not proposed earlier. Third, the proposed method can predict new combinations of labels, while NN-MLM is limited to the label sets seen in the training data. These are contributions related to the IDW-based MLM method. In addition, we have included many smaller novel methodological improvements and proposals related to the MLM based multi-label classification. We proposed: $i)$ distance regression model-based instance-wise thresholding (local RCut); $ii)$ we proposed to use a reference point corresponding to the smallest as BAN for LLS-MLM; $iii)$ we provide theoretical insights for NN-MLM in MLC; $iv)$ we compare MLM-based methods for MLC in detail; $v)$ we provide time complexity analysis for distance regression based methods; $vi)$ we provide insights to assess uncertainty with distance regression based methods. In addition, we provide a comprehensive literature review and comparison of the state-of-the-art MLC methods. To summarize, our main contributions are: 
\begin{itemize}
\item A novel method integration of IDW and MLM for MLC
\item Theoretical results for MLM in MLC problems
\item Methodological improvements for MLM
\item A comprehensive literature review of MLC
\item A comprehensive experimental comparison of the state-of-the-art MLC
\end{itemize}

% Structure of the paper
The remainder of the paper is structured as follows. First, in Section \ref{sec:relatedwork}, we review MLC methods and experimental comparisons and summarize the articles most closely related to our work. Then, in Section \ref{sec:background}, the notation and basic methodology are introduced. In addition, we present theoretical results for MLM. Then, the main proposal of the paper, ML-MLM, is introduced in Section \ref{sec:idmc}. This is followed by the experimental part of the paper in Section \ref{sec:results}. Finally, Sections \ref{sec:discussion} and \ref{sec:conclusion} discuss and conclude the paper.

\section{Background}
\label{sec:relatedwork}

In this section, we provide the necessary background materials for the article by first reviewing the most central MLC methodology in Section \ref{subsec:review}. Following that, in Section \ref{subsec:relateddistance}, we review the most closely related articles concerning our primary contribution. Finally, in Section \ref{subsec:exp}, we summarize the experimental results of MLC to provide context for our results.

%In this Section, we provide necessary background materials for the article by first reviewing the most central MLC methodology in Subsection \ref{subsec:review}. Next, in Section \ref{subsec:relateddistance}, we review the most closely related papers concerning our main proposal. Finally, in Subsection \ref{subsec:exp}, we review experimental landscape of MLC.

\subsection{A brief review of multi-label classification}
\label{subsec:review}

The most straightforward approach to tackle multi-label classification (MLC) problems is the so-called binary relevance (BR) problem transformation method \cite{boutell2004learning,tsoumakas2007multi}. In the BR approach, models are trained for each label separately using the one-vs-all approach. The prediction is then directly determined by these label-wise SLC model predictions. For the BR, there are well-known shortcomings. For a large number of labels, BR, and also other problem transformation methods, can have a high computational complexity. In addition, it is assumed in the BR that the information of label dependencies is not needed in the training, i.e., each predicted label is solely determined by the input data. Despite its shortcomings, the BR is an important methodological baseline for the MLC.

As depicted in \cite{tsoumakas2007multi} and Section \ref{sec1}, the MLC techniques can be basically divided into \emph{problem transformation (PT)} and \emph{algorithm adaptation (AA)} methods.
%, where the former refers to casting the MLC problem as one or many single label problem(s) and the latter refers to direct modification of a learning algorithm. 
In addition to BR, another classical example of PT is the label powerset (LP) \cite{tsoumakas2010random}, where each distinct combination of labels in the training set is treated as a single and different class. In fact, there exist many open source software packages based on PT techniques that provide implementations of MLC algorithms by interfacing with standard single-label methods \cite{read2016meka,szymanski2019scikit,wever2021automl}. {The CC method \cite{read2011classifier} is based on using the predictions of the BR model to extend the feature space with a chain structure. In this chain structure, an BR model is trained to predict a target label with a full set of features and a subset of predicted labels (at least the target label is excluded).}

%From Tommi 
%\noindent NEW\newline
%As depicted in \cite{tsoumakas2007multi}, techniques for multi-label classification (MLC) can be basically divided into \emph{problem transformation (PT)} and \emph{algorithm adaptation (AA)} methods, where the former refers to casting the MLC problem as one or many single label problem(s) and the latter refers to direct modification of a learning algorithm. Classical examples of PT are the binary relevance, BR \cite{boutell2004learning}, where separate, independent problems are used for each label, and label powerset, LP \cite{tsoumakas2010random}, where each distinct combination of labels in the training set is treated as a different, single class. There exists also many open source software packages based on the PT techniques which provide implementations of MLC algorithms via interfacing with the single-label methods \cite{read2016meka,szymanski2019scikit,wever2021automl}.

According to \cite{zhang2013review}, the AA methods can be further divided into first- and second-order methods, where the classical k-nearest-neighbor (kNN) MLC extension, ML-kNN \cite{zhang2007ml}, is an example of the former, and the rank-SVM with a modified kernel \cite{elisseeff2001kernel}, of the latter. The first-order AA methods, also a straightforward extension of single-label multiclass classifiers, can produce by construction an output-vector reflecting class probabilities\textemdash such as shallow and deep neural networks with 1-of-$k$ class encoding\textemdash, but with a modified labeling strategy \cite{triguero2016labelling}. In the simplest case, one could just return all certain enough labels, which can be selected using a threshold \cite{elisseeff2001kernel,triguero2016labelling,wu2016ml}, using a complete rejection of the unreliable predictions \cite{pillai2013multi}, or via the estimated posterior probabilities, according to Bayes' rule, as in ML-kNN \cite{zhang2007ml}. {The BPNN method in \cite{zhang2006multilabel} is an adaptation of the backpropagation algorithm to optimize the weights of a neural network for the MLC.}

%  {The BPNN method \cite{zhang2006multilabel} is based on a conventional neural network architecture which is based on backpropagation to update its weights and issues predictions.}

In addition to the AA and PT, a third category of the MLC approaches can also be defined, the ensemble methods \cite{madjarov2012extensive}. These methods refer to different kinds of combinations of the common problem transformation and algorithm adaptation methods. A prime example is the RAKEL (RAndom k-labELsets) method proposed by Tsoumakas \textit{et al.} \cite{tsoumakas2007multi}. { The HOMER method \cite{tsoumakas2008effective} is based on using a type of construction with a label power set with a hierarchical structure of the label space. In the structure, a set of binary classifiers are trained with the divide-and-conquer paradigm to predict meta-labels (a label power set of a subset of labels). In the training phase, the hierarchical structure of the labels is obtained by k-means balanced clustering.} Ensembles for PT that better address the multiscale nature of MLC problems, taking into account label dependencies, are provided by ensemble classifier chains (ECC), i.e. cascade combinations of single label models \cite{read2011classifier,spyromitros2016multi,jun2019conditional}. These and other ensembles are based on a base classifier that can be trained for different subsets of labels \cite{rokach2014ensemble,wang2021active}, different training examples \cite{tsoumakas2008effective,kocev2013tree}, using different subsets of features \cite{pereira2018categorizing,ma2019multilabel,weng2020non}, or via combinations of these divisive strategies \cite{huang2016learning}. Features and/or labels could also be transformed into a lower dimension space, either for a single MLC model or within an ensemble \cite{kumar2019group,huang2019supervised,siblini2019review,wu2020joint}. There exist multiple ways to integrate different base classifiers in ensembles, such as direct mean output \cite{rokach2014ensemble}, selection and use of the most prominent \cite{zhu2023dynamic}, and the weighted stacking strategy \cite{xia2021multi}. {The RF-PCT %\cite{kocev2013tree} 
method \cite{kocev2013tree} uses a decision tree with a hierarchical clustering structure as a base classifier for the Random Forests (RF) ensemble approach. In RF, the ensemble is constructed by randomizing the learning process on both an observation and feature direction with bootstrapping and random feature selection, respectively. The clustering structure of RF-PCT is obtained by maximizing variance reduction in a top-to-bottom direction.} Note that the new proposed MLM classifier for MLC could also be used naturally as a base classifier in ensembles, as demonstrated in \cite{hamalainen2020problem}.

As depicted in \cite{karkkainen2019extreme,hamalainen2020minimal}, the MLM technique can be linked to random NN techniques, which for the MLC problems were addressed in \cite{chauhan2022randomized}. There, functional link networks and broad learning systems were used as NN architectures with two variants each. The final results after training were computed using a trained threshold function, similar to \cite{elisseeff2001kernel}. Comparison was made with 12 benchmark datasets against three other random basis techniques using five evaluation metrics (Hamming loss, one error, coverage, ranking loss, and average precision). The experiments demonstrated high-quality results and improved computational efficiency.

Many new methods and approaches for the MLC have been suggested in recent years. One popular area, similar to ML in general (see \cite{linja2022feature}), has been to use and develop additional feature selection (FS) methods, such as filter \cite{lee2013feature,qian2021ranking}, wrapper \cite{he2019joint}, or hybrid/embedded form \cite{sun2021feature}. One can also select both features and instances, for example, based on their dependency in the latent topic space \cite{ma2022topic,ma2022multilabel}. General reviews of FS for the MLC, as provided in \cite{spolaor2016systematic,kashef2018multilabel,pereira2018categorizing}, were summarized in \cite[Section 2.6]{linja2022feature}. In conclusion, $i)$ most of the FS methods were of filter type; $ii)$ FS methods for the MLC problems could be categorized according to four perspectives: label, search strategy, interaction with the learning algorithm and data format; and $iii)$ there exists a direct relation between filter, wrapper and embedded FS methods with the single and internal/external BR techniques for the MLC problems. For example, label-specific FS techniques were suggested in \cite{weng2020non,yu2022multi}. 
%\newline \textcolor{red}{NOTE: In the latter much of same datasets than with us!}\newline  
FS using iterative search strategies for MLC problems, such as evolutionary \cite{bayati2022mssl} or multi-objective optimization \cite{dong2020many,hashemi2021efficient} approaches, have also been proposed. The use of rough sets for the ranking and selection of features was suggested in \cite{qian2021label}.

In summary, a plethora of different methods have been proposed and experimented with for the MLC problems over the years, with novel components and/or integration and modification of existing techniques with heterogeneous data sources. The basic categories of PT and AA have evolved and hybridized. For instance, one can construct new labels \cite{spolaor2016systematic}, address semi-supervised scenarios with incomplete multi-label information \cite{tan2017semi}, address streaming data and online/incremental learning \cite{nguyen2019multi1,nguyen2019multi2}, apply active learning techniques with an incremental query of the most relevant instances to improve the model's performance \cite{nakano2020active}, and extract higher-level features \cite{bello2020deep,yun2021dual}.

\subsection{Distance regression and distance weighting schemes}\label{subsec:relateddistance}

%As shown in the previous subsection, there exists a vast amount of diverse research work. 
In this subsection, we will review related research more closely associated with our proposal. We consider closely related works, the ones that use distance regression-based construction for MLC. We also review the most relevant work in IDW.

%As shown in the previous Sections \ref{subsec:review} and \ref{subsec:exp}, there exists a vast amount of diverse research work, extensive empirical comparisons, and well-written recent reviews on the MLC topic \cite{bogatinovski2022comprehensive,tarekegn2021review}. Next, we will review related research more closely associated with our proposal. We consider closely related works the ones that use distance regression-based construction for MLC. We also review the most relevant works concerning inverse distance weighting.

Our proposed method uses a distance regression step from MLM \cite{de2015minimal}. Therefore, the most closely related work is \cite{hamalainen2021instance}, where MLM was adapted to the MLC problems with a simple nearest-neighbor heuristic. In this method, the multilateration problem is solved approximately to obtain a classification. This method differs from NN-MLM \cite{Mesquita2017} only with the label space assumptions. This difference is further considered in Proposition \ref{prop:limitations-nn-mlm} (Section \ref{sec:background}). Furthermore, in \cite{hamalainen2021instance}, a clustering-based NN-MLM approach was proposed for MLC to allow MLM scaling for large-scale data sets. The preliminary results there showed that the proposed methods outperformed ML-kNN in accuracy and Hamming loss metrics and that the performance was comparable to a random forest-based state-of-the-art method. In this paper, we will not consider this clustering-based NN-MLM approach for MLC further, since we focus on the evaluation of the proposed method thoroughly with several MLC metrics for small to moderate-sized data sets. Note that the state-of-the-art is different compared to those for large-scale MLC problems. Most of the recent state-of-the-art methods for large-scale MLC problems (including extreme MLC) are based on deep learning \cite{liu2021emerging, Bhatia16}. However, some clustering-based methods, such as SLEEC \cite{bhatia2015sparse}, still perform well in comparisons \cite{zhang2018deep}.

Due to its simplicity, IDW has been used in several data-driven methods and applications \cite{Joseph2011idw,Chen2012idw,Chen2019idw,Cevik2020idw,Bemporad2020idw,Shi2020idw}. However, unlike our proposal, these works only use IDW in the feature space. In our proposal, we use IDW in the output space to add the ranking loss optimized regression layer to the method, which is used to form the local weighting of the output space vectors for classification purposes.

% IDW in various applications (more or less applied in the original form)
Joseph and Kang \cite{Joseph2011idw} integrated IDW with linear regression for improved prediction accuracy. The proposed approach was tailored for regression tasks with large-scale data sets and high-dimensional input spaces. The proposed approach was able to provide confidence intervals for the prediction. Shi \textit{et al.} \cite{Shi2020idw} adapted IDW to detect anomalies in distributed photovoltaic power stations. The IDW method was used nearly in the original form and with the power parameter selection $P = 2$ suggested in the literature (e.g. \cite{shepard1968two}). The main idea of the method is to predict the power station's output with IDW and then compare it to the actual value to determine via thresholding whether the station is working anomalously. Chen and Liu \cite{Chen2012idw} applied IDW to a spatial regression task (rainfall prediction) and studied the behavior of the power and radius of influence parameters. The value of the power parameter was tuned via cross-validation, with optimal values in the interval $[0,5]$.

% IDW related method development
Chen \textit{et al.} \cite{Chen2019idw} integrated IDW with the mixed-effect model to construct a two-step method to handle missing data. This approach was further integrated with extreme gradient boosting to develop a robust method (missing data handling) for regression. The method was successfully applied to predict small concentrations of ambient particles at the ground level for high levels of missing data (nearly $90 \%$). In \cite{Cevik2020idw} Cevik proposed an IDW-based local descriptor for facial recognition. Similarly to the goals in \cite{Chen2019idw}, the descriptor aimed to provide a more robust presentation of the feature space. More precisely, it aims to tackle rotation variances and noise effects of the feature space. The IDW approach was adjusted with an additional distance decay parameter for a better match on a local pixel region depending on the distances and pixel intensities. The proposed method was found to perform well in the classification accuracy compared to the state-of-the-art. Bemporad \cite{Bemporad2020idw} integrated IDW and radial basis functions (RBF) to develop an acquisition function based on that global optimization algorithm for improved probing of expensive objective function evaluations. Exclusively from a methodological point of view, this combination of IDW and RBF is somewhat close to the method proposed in this paper, since there are similarities between MLM and RBF. For example, the universal approximation proof for RBF can also be applied for MLM \cite{hamalainen2020minimal}.

\subsection{%Experimental landscape
{Summary of empirical results}}\label{subsec:exp}

Next, without being exhaustive, we summarize the most relevant experimental comparisons of different methods during the last decade %that can be directly linked 
in comparison 
to our own experiments as reported in Section \ref{sec:results}. The reviewed experiments took place since 2016 and are considered extensive based on the following two criteria: $i)$ at least ten datasets were used, with at least one of them with more than 10 000 instances; $ii)$ at least five other algorithms were included in the comparison.

%The one below much too simple and superseeded by many others...
%\cite{pakrashi2016benchmarking} provided a small comparison of popular MLC algorithms, available in the MULAN Java library, using ten datasets with the maximum of 5000 instances (corel5k). The straightforward Label Powerset (LP), Classifier Chains (CC), and Backpropagation for Multilabel Learning (BP-MLL) were concluded as the best methods. 

In \cite{wu2016ml}, the authors proposed a hierarchical tree-based ensemble with an SVM base classifier, referred as \emph{ML-Forest}. %,  for linear, radial basis, or polynomial kernel. 
The work provided a comparison with eight other state-of-the-art algorithms using five-fold cross-validation (CV) for the parameter tuning based on the Hamming loss. The experiments included 12 datasets, 
%Datasets: scene , Reuters(10), Reuters(21), Reuters(90), medical, genebase, ohsumed, emotions, tmc2007, bibtex, corel5k, yeast. 
with the maximum number of training instances c. 21 500 ('Tmc2007').
Based on nine bipartition-based quality metrics,  
%Hamming loss, Example-based accuracy, Example-based precision, example-based recall, example-based F1, subset accuracy, macro-precision, macro-recall 
the proposed technique, balanced clustering-based hierarchy of multi-label classifiers (HOMER, \cite{tsoumakas2008effective}), and the random forest of predictive clustering trees (RF-PCT, \cite{kocev2013tree}) provided the best performances. However, for the four ranking-based metrics, 
%One-error, Coverage, Ranking loss, Average precision
the two stage architecture (TSA, \cite{madjarov2012two}) and the BR method from \cite{tsoumakas2007multi} showed slightly better performances, compared to the RF-PCT and ML-Forest methods. Concerning training time, RF-PCT (with the number of models fixed to 50) was the fastest among the best performers.

%In \cite{spolaor2016systematic} experiments with ten MLC datasets
%(Cal500, Corel5k, Corel16k001, Emotions, Fapesp, Genbasen, Llog-fn, Magtag5k, Scene, Yeast) with at most c. 14 000 examples (Corel16k001) concluded competitive performance based on F-measure, Hamming loss, Accuracy, and Micro-averaged F-measure F$_b$.

%In \cite{zhang2020elm} only 6 datasets, all less than 2500 examples -> Omit. A sufficient performance with a random basis method ELM reported .

%Below 10+ datasets but the max. number of examples reduced to 5 000. So not included here...
%\cite{he2019joint} proposed to use sparse feature representation with possible missing label information. 
% Same true for \cite{wu2020joint}, max. nbr of insts 5000

%Primarily good-lookings tests in \cite{xia2021multi}, BUT, max nbr of examples 2500 and not RF-PCT included in comparisons. In preprint a dataset table included but not in the final article!

%13 datasets,  but max. with 5000 examples \cite{sun2021feature}

%10 datasets, max. 2417 examples in \cite{qian2021ranking}; and the comparison there focus on FS not quality of MLC as is

In \cite{wang2021active}, two variants of a label selection strategy were developed to obtain balanced label subsets in ensemble learning, using SVM with the RBF kernel as the base classifier with a multi-class classification strategy of one-versus-all. The extensive experiments and the related comparison with other methods were based on 30 datasets with at most c. 44 000 examples ('Mediamill') and even almost 13 000 features ('EukaryoteGo'). The two proposed algorithms were compared with 12 other MLC-related methods using five repetitions of ten-fold CV with four quality criteria: label-based accuracy (1-Hamming loss), example-based F-measure and label-based macro/micro F-measures. In conclusion, all methods performed equally and satisfactorily with respect to the accuracy, but the proposed methods outperformed traditional PT methods, providing quality similar to that of the state-of-the-art ensemble methods for the example- and label-based F-measures. Interestingly, a couple of datasets ('CS' and 'Chess' with only binary features) were encountered where none of the methods provided satisfactory results. Moreover, it was also concluded that the micro F-measure is more stable than the macro F-measure to evaluate and compare different methods.

%\newline
%\textcolor{red}{NOTE: Many same datasets than in our comparison! Remember to check our results against these!}
% JOONAS: Not directly comparable due to 5-CV cross-validation
%\noindent \textcolor{red}{NOTE: Names of quality metrics not fixed - check and unify wording!}

While introducing and experimenting with the label-specific FS method \cite{yu2022multi}, a comparison with 16 benchmark datasets and seven other reference methods was provided, based on Hamming loss, one-error, ranking loss, average precision and macro-averaging AUC quality metrics. The estimates compared and their variability were computed using a five-fold CV. The experiments were concluded in favor of the proposed WRAP (WRAPping multi-label classification with label-specific features generation) approach.

%\newline 
%\textcolor{red}{NOTE: Many same datasets than with us!}
%
% JOONAS: Not directly comparable due to 5-CV cross-validation

%CAL500, emotions, birds, medical, language log, enron, scene, yeast, slashdot, corel5k, bibtex, corel16k, delicious, eurlex-sm, tmc2007, mediamill

A recent very thorough comparison of 26 methods using 42 benchmark datasets was reported in \cite{bogatinovski2022comprehensive}. 
%, with four sets of altogether 20 evaluation criteria (example- and label-based, ranking-based, and efficiency), 
The predictive performance of the methods was assessed using 20 evaluation metrics. The comparison concluded that the tree-based ensembles based on random forests, especially the RF-PCT and RFDTBR (Binary Relevance with Random Forest of Decision Trees) problem transformation method {(in this paper we refer to this as BR-RF)}, were always among the best performing strategies. The difficulty of nominating the best method based on multiple evaluation criteria and perspectives was emphasized. This conclusion is also usually given in FS reviews \cite{linja2022feature}.
%The predictive performance of the methods was assessed using four evaluation metrics: Hamming Loss, F1 example-based, Micro precision and AUPRC (Micro-averaged area under the precision-recall curve). This comparison concluded that the tree-based ensembles based on random forests, especially RF-PCT and RFDTBR (Binary Relevance with Random Forest of Decision Trees) problem transformation method were always among the best-performing methods. This comparison also concluded the difficulty (impossibility) of nominating the one best method based on multiple evaluation criteria and perspectives. Similar conclusion is also typically given in FS reviews \cite{linja2022feature}.

The most recent experiments were reported in \cite{zhu2023dynamic}, where an approach (MLDE) was proposed for dynamic selection and combination of base classifiers of an ensemble, by assessing both the accuracy and the ranking loss of the ones generated on the fly. The comparison presented was based on 24 datasets (all from \url{http://www.uco.es/kdis/mllresources/}) and used 9 other baseline methods: RAKEL \cite{tsoumakas2007multi}, ML-kNN \cite{zhang2007ml}, ECC \cite{read2011classifier}, 
%UPS: Same ref two times -> \cite{Read2011cc} omitted
ACkELo \cite{wang2021active}, MLWSE  \cite{xia2021multi}, LF-LELC \cite{zhang2021multi}, ELIFT \cite{wei2018ensemble}, BOOMER  \cite{rapp2020learning}, and DECC \cite{trajdos2019dynamic}. Logistic regression, decision tree, Bayes network, and SVM were used as base classifiers and accuracy, F1-measure, Hamming loss, one error, and average precision as the quality metrics. Based on the five-times repeated 2/3-1/3 division of the data into training and test sets, it was concluded that the proposed MLDE method always performed as one of the best.
%, superior performance of the proposed method was reported. 
%However, details about the division of data into training and test %division sets was not explained thoroughly. So, the result may be based on using the whole dataset as training data, therefore not comparing the generalization capability of the methods as estimated using CV techniques, as in the other comprehensive comparisons summarized above.
% UPS, my mistake "Each benchmark dataset is divided into two parts: training and testing datasets, using the iterative stratified method [45] with the partitions of 67% and 33%, respectively. To obtain more reliable results, this process is repeated 5 times

{
\section{%Methodological background
Multi-label minimal learning machine} \label{sec:background}

%\paragraph{Problem and notation}

This section is organized as follows. In Section \ref{sec:mlm_formulae}, we first provide the basic formulation of MLM and the theoretical results of the NN-MLM approximation. In Section \ref{sec:main_proposal}, we derive our primary proposal based on the MLM's distance regression mapping and the inverse distance weighting scheme. Finally, in Section \ref{subsec:idwmlm_tuning}, we derive ranking loss based closed form solution for the main proposal's model selection.

%{
%Explain the structure of this Section as soon as we have fixed the final organization of contents.
%}

\subsection{Basic formulation with an approximation result} \label{sec:mlm_formulae}
}
%We are interested in MLC tasks, which 
The MLC task 
corresponds to a supervised learning problems where inputs can be associated with multiple class labels. Formally, we are given a dataset $\mathcal{D}=\{(\mathbf{x}_i, \mathbf{y}_i)\}_{i=1}^N$ of input-output pairs $(\mathbf{x}_i, \mathbf{y}_i) \in \mathbb{R}^M \times \mathbb{Y}$ representing i.i.d. samples from an unknown joint distribution $p^\star(\mathbf{x}, \mathbf{y})$. The set $\mathbb{Y}$ denotes the universe of all possible multi-label assignments, which we denote as $\mathbb{Y}=\{0, 1\}^L$ --- $L$ is the total number of classes, and the $c$-th component of an output $\mathbf{y}$ is set to 1 if the corresponding example belongs to class $c$ and $0$ otherwise. We want to find a mapping/model $h: \mathbb{R}^M \rightarrow 
\mathbb{Y}$ such that $h(\mathbf{x}) = \mathbf{y}$ with high probability for any $(\mathbf{x}, \mathbf{y}) \sim p^{\star}$.

%\paragraph{Minimal learning machine} 
Minimal learning machine (MLM, \cite{de2015minimal}) is a supervised method whose training consists of fitting a multiresponse linear regression model between distances computed %from 
in 
the input and output spaces. Predictions for new incoming inputs are achieved by estimating distances using the underlying linear model, followed by a search procedure in the space of possible outputs.

Formally, let $\mathcal{X}=\{\mathbf{x}_i\}_{i=1}^N$ be a set of input vectors $\mathbf{x}_i \in \mathbb{R}^M$ and $\mathcal{Y}=\{\mathbf{y}_i\}_{i=1}^N$ their corresponding outputs $\mathbf{y}_i$. Also, let $\mathcal{R}=\{\mathbf{r}_i\}_{i=1}^K \subset \mathcal{X}$ be a $K$-length subset of $\mathcal{X}$ with the corresponding outputs $\mathcal{T}=\{\mathbf{t}_i\}_{i=1}^K \subset \mathcal{Y}$. The prediction $\hat{\mathbf{y}}$ for an input $\mathbf{x}$ obtained from a MLM with parameters $\mathbf{B} \in \mathbb{R}^{K \times K}$ is given by    
\begin{align}\label{eq:mlm}
 		\hat{\mathbf{y}} = h(\mathbf{x}) =  \argmin_{\mathbf{y} \in \mathbb{Y}} \left\{ J(\mathbf{y}) = \sum_{k=1}^K \left[\|\mathbf{y} - \mathbf{t}_k\|^2 -  \left(\sum_{i=1}^K B_{i,k} \|\mathbf{x} -\mathbf{r}_i \|\right)^2 \right]^2 \right\}.
\end{align}
The term $\sum_{i=1}^K B_{i,k} \|\mathbf{x} -\mathbf{r}_i \|$ corresponds to the estimate of the distance between $\mathbf{y}$ and $\mathbf{t}_k$ --- we also denote it by $\hat{\delta}_k$. Notably, \autoref{eq:mlm} can be viewed as a multilateration problem --- finding the coordinates of a query point from distance estimates to fixed (anchor) points.

Before solving the optimization problem described in \autoref{eq:mlm}, we need to determine the parameters $\mathbf{B}$.
Let us define $\Dx \in \mathbb{R}^{N \times K}$ as the pairwise Euclidean distance matrix between $\mathcal{X}$ and $\mathcal{R}$, that is, $D_{ij}= \|\mathbf{x}_i - \mathbf{r}_j\|$. Similarly, $\Dy \in \mathbb{R}^{N \times K}$ denotes the pairwise distance matrix between $\mathcal{Y}$ and $\mathcal{T}$. In its original formulation, MLM adopts $\hat{\mathbf{B}}= (\Dx^T\Dx)^{-1}\Dx^T\Dy$, which corresponds to the ordinary least-squares solution for a linear model between the input and output distances. 

For SLC tasks where classes are balanced, \cite{Mesquita2017} shows that the minimization in \autoref{eq:mlm} is equivalent to selecting the nearest output reference point (or class label) approximated with the distance regression model, i.e., we can simply compute
\begin{align}
    \hat{\mathbf{y}}=\mathbf{t}_{k^\star} \quad \text{ with } \quad k^\star = \argmin_{k=1,\dots,K} \left\{\hat{\delta}_k = \sum_{i=1}^K \hat{B}_{i,k} \|\mathbf{x} -\mathbf{r}_i \| \right\}.
\end{align}
This strategy was called nearest neighbor MLM (NN-MLM), which was proposed for multi-label settings in \cite{hamalainen2021instance}. %proposed applying the NN-MLM to multi-label settings. 
In this regard, Proposition 1 shows that this nearest neighbor procedure does not yield optimal solutions for MLC for general values of $\hat{\delta}_k$. However, the NN-MLM does correspond to the solution of the multilateration problem in \autoref{eq:mlm} when estimates of the output distance %estimates 
are accurate. This explains the high predictive performance of the NN-MLM for MLC reported in  \cite{hamalainen2021instance}. We provide a proof in \autoref{ap:proof}.

\begin{theorem}[Nearest neighbor MLM for multi-label learning] \label{prop:limitations-nn-mlm} Let $L \geq 2$ denote the number of classes in an MLC problem with $\mathbb{Y}=\{0,1\}^L$. 
Also, assume that the set of output reference points $\mathcal{T}$ contains all possible multi-label assignments. Then, the following holds:
\begin{enumerate}
    \item For arbitrary values of $\hat{\delta}_k$, NN-MLM does not always yield optimal solutions to the optimization problem in \autoref{eq:mlm}. 
    \item If $\hat{\delta}_k = \|\mathbf{y}' - \mathbf{t}_k\|$ for some $\mathbf{y}' \in \mathbb{Y}$, then NN-MLM returns an optimal solution to the minimization in \autoref{eq:mlm}.
\end{enumerate}
\end{theorem}

{
The problem of selecting the sets $\mathcal{T}$ and $\mathcal{R}$ is referred as reference point selection in the MLM %methods 
nomenclature. This problem is closely related to the selection of landmark points for the Nyström method \cite{hamalainen2020minimal}. In order to simplify experimentation and analysis, in this paper, we focus on using a full set of output space reference points $\mathcal{T}$, and all distinct points from input space as input space reference points $\mathcal{R}$. Note that there are several proposals for the reference point selection in the MLM literature \cite{RSesann2018,florencio2018,Maia2018,hamalainen2020minimal} to improve the model performance.
}

%\section{Multi-label minimal learning machine}
\subsection{Multi-label algorithm} \label{sec:main_proposal}
\label{sec:idmc}

%{
%Joonas: Check that the notations and derivations below do not overlap those which were already made in Section 3.1. For me it seems that some issues and notations and formulations are done twice, first in 3.1 and then here again.
%}
% Checked those, for me those seems correct. There is some rementioning of variables but not overlapping -Joonas

In practice, meeting the requirements that ensure the optimality of the NN procedure (Proposition 1) is often hard. In addition, the greedy nearest label choice limits the usage of the NN-MLM only for a seen set of labels, i.e., predictions are limited to those cases which have been observed during training. Thus, here we take an alternative choice and leverage the full set of labels, going from $1$-NN to $N$-NN. To do so, we need to define a way to combine the labels. In summary, our solution relies on computing a convex combination where importance weights are given as a function of distances in the label space (estimated using the values $\hat{\delta}$'s). Motivated by the success of nearest neighbors in multi-label settings \cite{hamalainen2021instance}, we target at weighting schemes where the first nearest neighbor should have higher importance than the second one, and so on.

A natural way to implement this is via IDW \cite{shepard1968two}. This scheme allows us to control the weighting of the convex combination of label sets with a single parameter. The power parameter. In the following, we formulate this idea and introduce ML-MLM (multi-label MLM) --- a new distance-based method for MLC which combines the distance regression step from the standard MLM with IDW.

%\subsection{Formulation}

Similarly to MLM, ML-MLM assumes the existence of a linear mapping between the distances computed from the input and output/label data spaces. Under the least-squares criterion, we can obtain a closed-form estimate of the linear coefficients. Consider the sets $\mathcal{X}$, $\mathcal{Y}$, $\mathcal{R}$, and $\mathcal{T}$ as in the original MLM, except for the fact that $\mathcal{T}=
\mathcal{Y}$ --- all observed targets are also output reference points. Also, let $\Dx \in \mathbb{R}_+^{N \times K}$ be the pairwise $\ell_2$-distance matrix between the $\mathcal{X}$ and $\mathcal{R}$, whereas $\Dy \in \mathbb{R}_+^{N \times N}$ is the distance matrix between the elements of $\mathcal{Y}$.
Training for ML-MLM consists of simply computing
$\hat{\mathbf{B}}= (\Dx^T\Dx)^{-1}\Dx^T\Dy
$, where $\hat{\mathbf{B}} \in \mathbb{R}^{K \times N}$ denotes the model parameters. 

\begin{figure}[h]
  \centering
    \includegraphics[width=\textwidth]{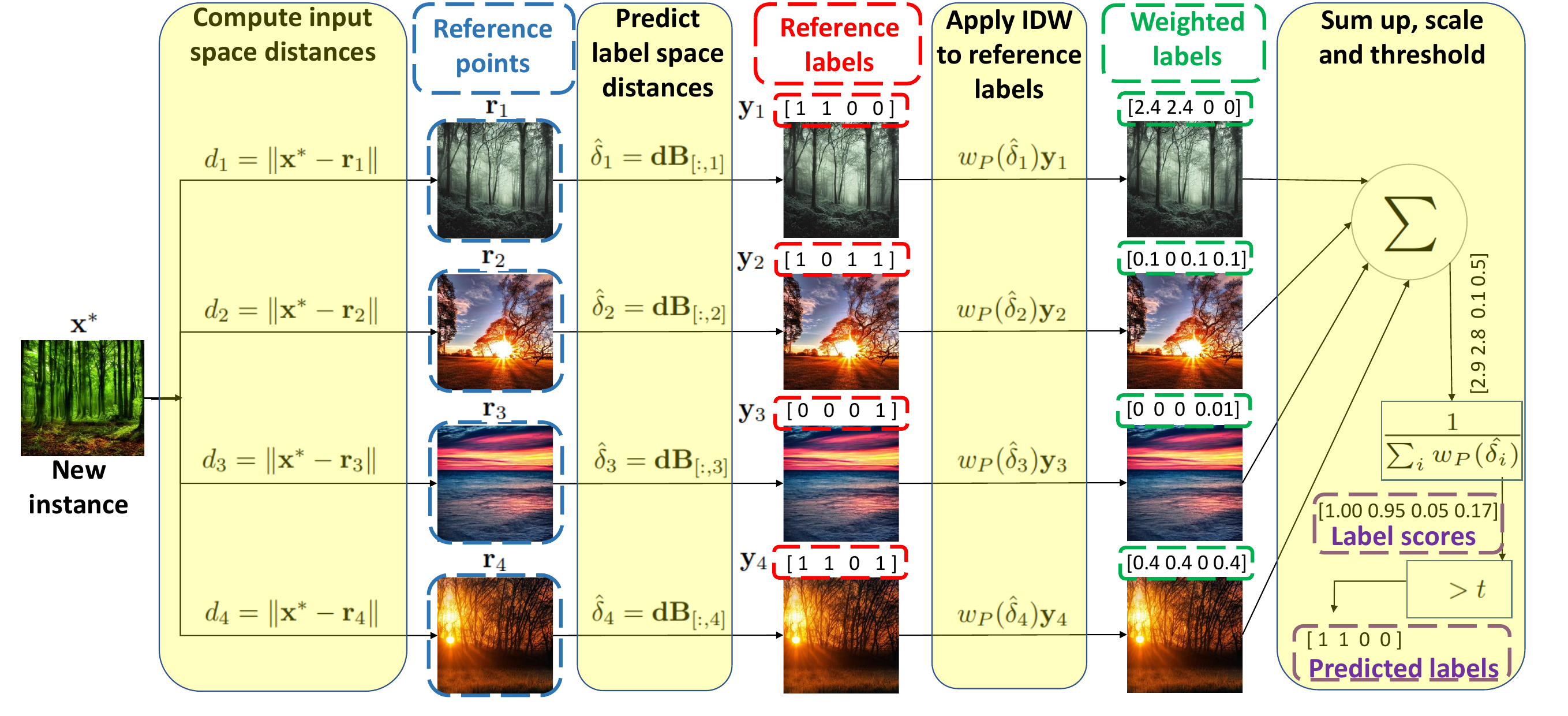}
    \caption{The main components of the ML-MLM's prediction in action are illustrated for a model trained on a toy dataset. The dataset consists of four instances associated with four unique label sets. A distance regression model $\mathbf{B}$ has been trained with four input space reference points $\{\mathbf{r}_i\}_{i=1}^4$ with corresponding label space vectors $\{\mathbf{y}_i\}_{i=1}^4$ (treated as output space reference points regarding the MLM context). The IDW weighting is computed with $P = 2$. The sample images here were created by Stable Diffusion.}
	\label{fig:mlc_demo}
\end{figure}

To obtain a multi-label prediction for a new input $\mathbf{x}$, ML-MLM first applies the distance regression model to get the distance estimates $\hat{\bm{\delta}} = [\hat{\delta}_1,\cdots,\hat{\delta}_N]$ with respect to the observed set of multi-label targets $\mathcal{Y}$, i.e.,
\begin{equation}
\hat{\bm{\delta}} = \left[\left \| \mathbf{x} -  \mathbf{r}_1 \right \|_2, \cdots, \left \| \mathbf{x} -  \mathbf{r}_K \right \|_2 \right] \hat{\mathbf{B}},
\end{equation}
where $\mathbf{r}_i$ denotes the $i$-th element of $\mathcal{R}$. We then use these distance estimates to define the importance weight \cite{shepard1968two} for each target in $\mathcal{Y}$, and compute a weighted average over the observed multi-label targets:
\begin{align}
\bar{\mathbf{y}} = \frac{1}{Z}\sum_{i=1}^N w_P(\hat{\delta}_i) \mathbf{y}_i, \quad \text{ with } \quad w_P(\hat{\delta}_i) = \begin{cases}
                                    \hat{\delta}_i^{-P},&           \text{if } \hat{\delta}_i > 0,\\
                            1,    & \text{if } \hat{\delta}_i = 0,
\end{cases}   
\end{align}
where $Z=\sum_{i=1}^N w_P(\hat{\delta}_i)$ and $0 < P \in \mathbb{R}$ is a hyper-parameter. In particular, the variable $P$ controls how fast the importance weight decays with the distance $\hat{\delta}_i$. Finally, we achieve the multi-label prediction $\hat{\mathbf{y}}=[\hat{y}_1, \ldots, \hat{y}_L] \in \{0, 1\}^L$ by thresholding the values of $\bar{\mathbf{y}}$:
\begin{align}
    \hat{y}_c = \mathbbm{1}[\bar{y}_c > t] \quad \quad \forall c=1, \dots, L,
\end{align}
where $\mathbbm{1}[\cdot]$ denotes the indicator function that returns 1 if its argument is true and 0 otherwise, and $t$ is a hyper-parameter. Algorithms \ref{alg:mlmtrain} and \ref{alg:idwmlmpred} summarize the ML-MLM's training and prediction procedures. In Figure \ref{fig:mlc_demo}, the prediction workflow of Algorithm \ref{alg:idwmlmpred} is illustrated with a toy dataset.

\begin{algorithm}
%\caption{MLM training}
\label{alg:mlmtrain}
\caption{Distance regression (training)}\label{alg:mlmtrain}
\begin{algorithmic}[1]
\Require 
Input data $\mathcal{X}$, output labels $\mathcal{Y}$, input space reference points $\mathcal{R}$
\Ensure Distance regression model $\mathbf{\hat{B}}$
\State $\Dx \gets$ compute $N \times K$ input space distance matrix between $\mathcal{X}$ and $\mathcal{R}$ 
\State $\Dy \gets$ compute $N \times N$ distance matrix for $\mathcal{Y}$
\State $\hat{\mathbf{B}} \gets $ solve $(\Dx^T\Dx)^{-1}\Dx^T\Dy$
\end{algorithmic}
\end{algorithm}

\begin{algorithm}
\caption{ML-MLM prediction}\label{alg:idwmlmpred}
\begin{algorithmic}[1]
\Require Input vector $\mathbf{x}^*$, distance regression model $\mathbf{\hat{B}}$, reference points $\mathcal{R}$, output labels $\mathcal{Y}$, power parameter $P$, classification threshold $t$
\Ensure label vector $\mathbf{y}^*$
\State $\mathbf{d} \gets$ $[~ \|\mathbf{x}^*-\mathbf{r}_1 \|, ..., \| \mathbf{x}^*-\mathbf{r}_K \| ~]$ ~~~~~~~~~~~~// \text{Compute input space distances}
\State  $\bm{\hat{\delta}} \gets$ $\mathbf{d} \mathbf{\hat{B}}$ ~~~~~~~~~~~~~~~~~~~~~~~~~~~~~~~~~~~~~~~~~~~~// \text{Predict output space distances}
\State $\bar{\mathbf{y}} \gets \sum_i w_P(\hat{\delta}_{i}) \mathbf{y}_i / Z$ ~~~~~~~~~~~~~~~~~~~~~~~~~~~~~~~~// \text{Compute label scores (Eq. 4)}
\State $\mathbf{y^*} \gets $ $\mathbbm{1}[\bar{\mathbf{y}} > t]$ ~~~~~~~~~~~~~~~~~~~~~~~~~~~~~~~~~~~~~~~~// \text{Threshold relevant labels}
\end{algorithmic}
\end{algorithm}

\subsection{Model selection}\label{subsec:idwmlm_tuning} 
We now discuss how we tune the hyper-parameters $P$ (power) and $t$ (global class threshold). Regarding the selection of $P$, we employ a broadly used evaluation metric for multi-label problems: \textsc{Ranking Loss} \cite{zhang2013review}. The \textsc{Ranking Loss} is an appropriate choice for methods that provide label scores as the ML-MLM since it measures the overall quality of the ranking. We combine this loss with a CV scheme, or more specifically, leave-one-out CV (LOOCV). Notably, due to the linear nature of the ML-MLM distance regression step, we can leverage the prediction sum of squares (PRESS) statistic \cite{Allen1974} to compute the LOO error in a closed form. As a result, we only need to run the MLM's training algorithm once.

% LOOCV formulation for ranking loss
In Algorithm \ref{alg:mlmtrain}, the coefficient matrix $\mathbf{\hat{B}}$ can be solved with the Moore-Penrose pseudoinverse. For $\mathbf{U} = \Dx^T \Dx + \alpha \mathbf{I}$, we have $\mathbf{\hat{B}} = \mathbf{U}^{\dagger} \Dx^T \Dy$. For each training set observation $i \in \{1,...,N \}$, we can determine the corresponding out-of-sample distance prediction $\mathbf{\hat{\bm{\delta}}}_i^{LOO}$ as
%$\mathbf{d_{y^*}}$
\begin{equation}
\label{eq:loocv_dists}
\mathbf{\hat{\bm{\delta}}}_i^{LOO} =  \frac{\hat{\mathbf{D}}_{\mathbf{y}(i,:)} - \mathbf{H}_{(i,i)} \mathbf{D}_{\mathbf{y}(i,:) }}{1 - \mathbf{H}_{(i,i)}},
\end{equation}
where a hat matrix $\mathbf{H} = \Dx \mathbf{U}^{\dagger} \Dx^T$ and $\hat{\mathbf{D}}_\mathbf{y} = \mathbf{H}_{} {\Dy}_{}$. Therefore, the LOOCV \textsc{Ranking Loss} (LRL) statistic for ML-MLM is given by 
\begin{equation}
\label{eq:loocv_ranking_loss}
\textsc{LRL} =  \frac{1}{N} \sum_{i = 1}^{N} \frac{\vert \{ (j,k) ~\vert~ \bar{\mathbf{y}}_{i(j)} < \bar{\mathbf{y}}_{i(k)}, ~j \in g_i^+, ~ k \in g_i^0  \}\vert}{ \vert g_i^+ \vert\vert  g_i^0 \vert},
\end{equation}
%\begin{equation}
%\label{eq:loocv_ranking_loss}
%\textsc{LRL} =  \frac{1}{N} \sum_{i = 1}^{N} \frac{\vert \{ (j,k) : \mathbf{\tilde{z}}_{i(j)} < \mathbf{\tilde{z}}_{i(k)}, ~\mathbf{y}_{i(j)} = 1, ~ \mathbf{y}_{i(k)} = 0 \} \vert }{\vert \{ j: \mathbf{y}_{i(j)} = 1 \} \vert\vert { \{ k: \mathbf{y}_{i(k)} = 0 \}\vert}},
%\end{equation}
where $\bar{\mathbf{y}}_i = \sum_i w_P(\mathbf{\hat{\bm{\delta}}}_i^{LOO}) \mathbf{y}_i / Z$ is a score vector determined by a power parameter value and an out-of-sample distance prediction, $g_i^+ = \{ j: \mathbf{g}_{i(j)} = 1 \}$ is a set of label indices referring to relevant ground truth labels for an instance $i$, and, correspondingly, $g_i^0 = \{ j: \mathbf{g}_{i(j)} = 0 \}$ is a set of label indices referring to irrelevant ground ground truth labels for an instance $i$. Hence, the objective of ML-MLM training is to find a $P^* =  \argmin_P \textsc{LRL}$. The exploitation of Allen's PRESS formulation in ML-MLM differs from a conventional use case where the mean squared error calculation for each hyper-parameter value requires solving the least-squares problem each time again \cite{Eirola2015elmloo,Alencar2016elmloo}. In the ML-MLM training, the least-squares problem has to be solved only once for tuning the hyper-parameter. This difference is even more significant if we use a naive LOOCV implementation where the least-squares problem is solved $N$ times for each hyper-parameter value.

% Label cardinality based global thresholding  
Thresholding for the MLC models can generally be performed with label-wise or global thresholding. In \cite{alotaibi2021multi}, the authors concluded that there is no significant difference between the multiple and single thresholding approaches. The global threshold can be selected in an intuitive way by matching prediction's label cardinality \cite{tsoumakas2007multi} to a ground truth label cardinality.

After finding $P^*$, thresholding $t$ can be tuned for ${{\mathcal{Y}}_t} = \mathbbm{1}[\bar{\mathcal{Y}} > t]$, where $\bar{\mathcal{Y}} = \{\bar{\mathbf{y}}_i\}_{i=1}^N$ so that \begin{equation}
\label{eq:card_t}
\textsc{Card}({\mathcal{Y}_t}) = \textsc{Card}(\mathcal{Y}),
\end{equation}
 where $\textsc{Card}(\cdot)$ computes label cardinality (see Eq. \eqref{eq:label_cardinality} for the \textsc{Card} measure) for a given set of label vectors.

\section{Experiments and Results} \label{sec:results}

% Outline of the experimental comparison
In this section, we focus on comparing the proposed method (ML-MLM), three other distance regression-based (DRB) methods, and {13} reference methods with ten benchmark datasets. {Since we do not have the possibility to obtain ranking scores for all methods, we have different sets of methods for ranking-based and bipartition-based comparisons. For the ranking-based comparison, we compare three DRB methods and four reference methods. For the bipartition-based comparison, we compare all four DRB methods and 13 reference methods.} Implementations of the new methods are available in \url{https://github.com/jookriha/ml-mlm}. In our experiments, we focus on comparing the proposed method with a state-of-the-art method, RF-PCT, which was concluded to be among the best performers in the extensive comparison by Bogatinovski \textit{et al.} \cite{bogatinovski2022comprehensive} (see Section \ref{sec:relatedwork}). Regarding DRB methods, we are especially interested in comparing ML-MLM and LLS-MLM. In addition, we analyze the uncertainty of the proposal's prediction and its time complexity. In Section \ref{expset}, the experimental setup is described. In Section \ref{sec:res}, the results of the experimental comparison are given with four ranking-based metrics and four bipartition-based metrics. In Section \ref{sec:interpretation}, we demonstrate how the uncertainty of the prediction can be assessed. Finally, in Section \ref{sec:cost}, we analyze and compare the computational costs.

\subsection{Experimental Setup} \label{expset}
%
% Introducing metrics for the MLC dataset characterization and the benchmark datasets via these metrics
%
In MLC, datasets are often characterized by label cardinality and label density metrics \cite{tsoumakas2007multi}, which summarize how many labels are relevant per instance on average. The label cardinality is defined as
\begin{equation}
\label{eq:label_cardinality}
\textsc{Card} =  \frac{1}{N} \sum_{i = 1 }^{N} \mathbf{1}^T\mathbf{y}_i,
\end{equation}
where $\mathbf{1} \in \{1\}^L$. The label density is given by
\begin{equation}
\label{eq:label_density}
\textsc{Dens} =  \frac{1}{N} \sum_{i = 1 }^{N}  \frac{\mathbf{1}^T\mathbf{y}_i}{L}.
\end{equation}
The datasets used in the experiments are summarized in Table \ref{tab:data}, where the \textsc{Card} and \textsc{Dens} metrics have been computed for the union of the training and test set label vectors. In addition to these metrics and common data characteristics, we computed the number of unique label vectors for the union of the training and test sets, denoted as $L_u$, and the number of distinct combinations of labels (\#novel label vectors) in the test set, denoted as $L_n$. All datasets are available in \url{http://mulan.sourceforge.net/datasets-mlc.html} (the Mulan repository). %\footnote{\url{http://mulan.sourceforge.net/datasets-mlc.html}}.

% Metrics
The relevant background to follow and understand the empirical work in relation to MLC problems is to select and understand the quality metrics that are used in evaluating and comparing different methods. Clearly, the solution of an MLC problem itself is of multi-criteria nature because the simplest model, which is the fastest to train and provides the best performance by means of all possible metrics, should be obtained. This landscape was thoroughly evaluated in \cite{pereira2018correlation}, and the most relevant metrics used in our and others' experiments are summarized in Appendix \ref{app:metrics}. Implementations of the metrics are available in \url{https://github.com/jookriha/ml-mlm}.

% About the datasets
Since we compare our results to the ones given in \cite{madjarov2012extensive} for the ranking-based metrics, the selected datasets are the same, with the expectation that we excluded the Bookmarks dataset, since some of the reference results for it were missing. Moreover, for this dataset, training BR-MLM with a nearly full set of reference points would have been very expensive. Note that the Scene dataset is close to being a single-label classification problem, since label cardinality is $1.074$. Therefore, its relevance for MLC comparisons is somewhat questionable.

%
% Summary of the methods and evaluation metrics
%

%For the comparison, we computed the results for four DRB methods: MLM with Binary Relevance (BR-MLM), MLM with Localization Linear System (LLS-MLM, \cite{hamalainen2020minimal}), \mname, and multi-label adaptation of NN-MLM \cite{hamalainen2021instance}. Note that for NN-MLM, it is not possible to get the ranking, therefore, we computed only bipartition-based metrics for it.

For the comparison, we computed the results for the following four DRB methods: 
\begin{itemize}
\item \textbf{BR-MLM} uses cubic equations MLM (C-MLM) \cite{Mesquita2017} as a base model for the BR approach. In C-MLM, a closed-form solution for a label-wise multilateration problem is used. Each C-MLM model outputs directly a real-valued score for label ranking and classification. Since we use the local RCut thresholding, this method does not require hyperparameter tuning. Integration of these three elements is proposed in this paper.
\item \textbf{LLS-MLM} solves the MLM's multilateration problem via the localization linear system (LLS)  \cite{hamalainen2020minimal} approach, where predicted distances, benchmark anchor node (BAN), and output space reference points (all but not BAN) construct a linear system of equations. A solution to the linear system of equations with BAN transition outputs real-valued scores for all labels. These scores are directly used for label ranking and classification. We used the corresponding reference point to the smallest predicted distance as a BAN. Since we also use the local RCut thresholding for this method, hyper-parameter tuning is not needed.  Integration of these three elements is proposed in this paper.%The described BAN selection is proposed in this paper. 
\item \textbf{ML-MLM} is the main proposal of this paper. The method uses the IDW \cite{shepard1968two} approach to form a convex combination of the set of training label vectors. The two hyper-parameters, power parameter and global thresholding, can be selected straightforwardly. The out-of-sample distance predictions can be solved from closed-form solutions based on the ideas from the PRESS statistic \cite{Allen1974}. These predictions from the closed-form solution enable a one-shot training process using the LRL statistic for selecting both the power parameter and the label cardinality-based global thresholding parameter.
\item \textbf{NN-MLM} predicts a set of labels directly given by the corresponding reference point to the smallest predicted distance. This method was proposed in \cite{hamalainen2021instance}. Note that this method is a direct adaptation of the original NN-MLM \cite{Mesquita2017} to multi-label settings. Therefore, we will refer to it here with the same name as in the SL domain. The main difference is that for NN-MLM in MLC, there are no theoretical guarantees in solving the multilateration problem optimally (Proposition \ref{prop:limitations-nn-mlm}).
\end{itemize}

%Note that due to the linear distance regression model \ref{}, predicted distance to the output space reference points can be  

% Implementation details (common)
For all these four DRB methods, we use the same set of input space reference points $\mathbf{R}$. We selected all unique training instances as reference points $\mathbf{R}$. We kept the full set of output reference points since singularity issues concern only the inverse of the matrix $\Dx^T\Dx$. In \ref{eq:loocv_dists}, we fixed the $\alpha$ parameter as the lower 1000-quantile of the pairwise input space reference point distances and with an exclusion of a reference point's distance to itself (all the pairwise distances are nonzero). We observed that this heuristic improved unique solving of the inverse of $\Dx^T\Dx$. For LLS-MLM, ML-MLM, and NN-MLM, we always use the same distance regression model.

% Implementation details (method-wise)
In the LLS-MLM prediction algorithm (in \cite{hamalainen2020minimal}, see Algorithm 1), we select the benchmark-anchor-node (BAN) as an output space reference point corresponding to the smallest predicted distance. We conducted an extensive experimental study regarding the BAN selection. As a result, we observed the proposed selection strategy outperformed the originally proposed random-based selection in the MLC problems. Due to this BAN selection, the LLS-MLM prediction is deterministic in our experiments. Therefore, the four DRB methods used here are fully deterministic with respect to training and prediction. For the BR-MLM, we used the C-MLM \cite{Mesquita2017} for each SLC model, since we cannot get a label scoring/ranking from the nearest neighbor MLM \cite{Mesquita2017} method. For ML-MLM, we employed the LOOCV \textsc{Ranking Loss} statistic to tune the power parameter. Our preliminary exploration of \textsc{Ranking Loss} values as a function of the power parameter revealed the largest rate of change for lower-end values. Consequently, we searched for the power parameter value from the base of two $2^{s}$ values, where $s = {0,0.1,0.2,...,8}$.

% Thresholding with Local Rank Cut
For the BR-MLM and LLS-MLM methods, we observed that classification results were suboptimal with a fixed $0.5$ thresholding. Therefore, we proposed a new thresholding approach which we refer to as `local rank cut' (local RCut), where the labels corresponding to the highest scores $K_{cut}$ are labeled as relevant. Unlike the Rcut thresholding \cite{alotaibi2021multi}, in local RCut, the value $K_{cut}$ is given by the cardinality of the predicted set of labels of the NN-MLM model. For the BR-MLM and LLS-MLM methods, we omitted the label cardinality-based thresholding, since it would increase the training complexity significantly for these methods. The local RCut does not require any hyper-parameter tuning in the training for the BR-MLM and LLS-MLM methods.

% Thresholding for ML-MLM (label-cardinality based approach)
For ML-MLM, we used the label cardinality-based global thresholding, as described in Section \ref{subsec:idwmlm_tuning}. For ML-MLM, available out-of-sample distance predictions can be straightforwardly exploited for cardinality-based thresholding. Moreover, ranking performance for the \textsc{Ranking Loss} and \textsc{Coverage} metrics indicated that ML-MLM's thresholding should be tuned so that it aims to retrieve labels from deeper in the ranking than BR-MLM and LLS-MLM. We experimentally compared the label-cardinalities for the predicted set of labels between the cardinality-based and local RCut thresholding approaches with ML-MLM and observed that the average predicted cardinality of the label was lower for the local RCut thresholding. Hence, local RCut is a more conservative approach than cardinality-based thresholding.

{As recommended in \cite{madjarov2012extensive, bogatinovski2022comprehensive}, we compare our methods with BR \cite{tsoumakas2007multi, Zhang2018brreview}, CC \cite{read2011classifier,Read2021ccreview}, HOMER \cite{Tsoumakas2008homer}, RAKEL~\cite{tsoumakas2007multi}, BPNN~\cite{zhang2006multilabel, read2014bpnn} and RF-PCT \cite{Kocev2007rfpct}. Note that the RF-PCT method was identified as one of the best performing methods in both studies \cite{madjarov2012extensive,bogatinovski2022comprehensive} closely followed by BPNN in \cite{bogatinovski2022comprehensive} in AA methods. In addition, we also included the ML-kNN \cite{zhang2007ml} method for the comparison, because it belongs to the same group as the method proposed in the MLC method taxonomy \cite{tsoumakas2007multi, bogatinovski2022comprehensive}.}

%Results for BR, CC, and HOMER were produced with SVM as a base model \cite{madjarov2012extensive}.
{Following recommendations issued in \cite{madjarov2012extensive, bogatinovski2022comprehensive} to select the base model, BR and CC are produced with different base models including SVM and J48 decision tree learner~\cite{quinlan_induction_1986} namely hereafter BR-SVM, BR-J48, CC-SVM and CC-J48. To extend the results with BR, we also include two other complex base models, namely AdaBoost (BR-Ada) and Random Forest of Decision Trees (BR-RF). Following the same principles, the results for Rakel are produced with SVM (Rakel-SVC) and a stochastic gradient descent (Rakel-SGDC).}

{We used the results from \cite{madjarov2012extensive} for BR-SVM, CC-SCM, ML-kNN, HOMER, and RF-PCT. For BR-J48, CC-J48, BR-Ada, BR-RF, Rakel-SCV, Rakel-SGDC, ML-kNN (d), ECC-J48, EBR-J48, and BP-NN, we computed the classification results using the scikit-multilearn library's MEKA wrapper (\url{http://scikit.ml/userguide.html}) in a Python environment. For the latter list of classifiers, we used the default parameters suggested by MEKA. Note that we have results for the ML-kNN from \cite{madjarov2012extensive} and MEKA-based with the default parameter. We distinguish these with "(d)" denoting the default one. 
}

We used the given train-test splits for the datasets to compare our results with the results given in \cite{madjarov2012extensive}. In statistical testing, we used the Friedman test followed by the Nemeneyi post hoc test (if the null hypothesis was rejected) \cite{demvsar2006statistical}. The statistical significance level was set to $0.05$.

\begin{table}[t]
 \caption{Characteristics of datasets. On the top row of the table, $N$ is the number of instances in the training set, $N_{ts}$ is the number of instances in the test set, $M$ is the number of features, $L$ is the number of labels, $Q$ is the number of novel labels in the test set, \textsc{Card} is the label cardinality, \textsc{Dens} is the label density, $L_u$ is the number of unique label vectors, and $L_{n}$ is the number of novel label vectors in the test set.}\label{tab:data}
 \normalsize
 \centering
 \bgroup
 \def\arraystretch{1.2}%
 \begin{adjustbox}{max width=1\columnwidth}
 \begin{tabular}{lrrrrrrrrrrrr}
 \hline
 %\textbf{Dataset} & \textbf{\#samples} & \textbf{\#features} & $\bf{K}$ \\ \hline
\text{Dataset} & $N$ & $N_{ts}$ & $M$ & $L$ & $Q$ & \textsc{Card} & \textsc{Dens} & $L_u$ & $L_n$ \\ \hline
Medical & 333 & 645 & 1449 & 45 & 7 & 1.245 & 0.028 & 94 & 54\\
Emotions & 391 & 202 & 72 & 6 & 0 & 1.869 & 0.311 & 27 & 4\\
Enron & 1123 & 579 & 1001 & 53 & 1 & 3.378 & 0.064 & 753 & 225\\
Scene & 1211 & 1196 & 294 & 6 & 0 & 1.074 & 0.179 & 15 & 1\\
Yeast & 1500 & 917 & 103 & 14 & 0 & 4.237 & 0.303 & 198 & 38\\
Corel5k & 4500 & 500 & 499 & 374 & 3 & 3.522 & 0.009 & 3175 & 252\\
Bibtex & 4880 & 2515 & 1836 & 159 & 0 & 2.402 & 0.015 & 2856 & 836\\
Delicious & 12920 & 3185 & 500 & 983 & 0 & 19.020 & 0.019 & 15806 & 3095\\
Tmc2007 & 21519 & 7077 & 500 & 22 & 0 & 2.220 & 0.101 & 1172 & 0\\
Mediamill & 30993 & 12914 & 120 & 101 & 0 & 4.376 & 0.043 & 6555 & 2223\\
\hline 
\end{tabular}
\end{adjustbox}
\egroup
\end{table}

\subsection{Results} \label{sec:res}

In this section, we show experimental results for comparison. We analyze results for the ranking-based and bipartition-based metrics separately. Detailed results for each method, dataset and evaluation metric are given in eight tables, in Appendix \ref{app:ranking} and Appendix \ref{app:bipartition}, for the ranking-based and bipartition-based results, respectively. The results of these tables are visualized with the Nemeneyi diagrams in Figures \ref{fig:ranking} and \ref{fig:bipartition}. %evaluation, which is independent from the MLC thresholding.

\subsubsection{Ranking-based metrics} \label{sec:ranking}

\begin{figure}[t]
  %\centering
  \begin{subfigure}{0.45\textwidth}
    \includegraphics[width=\textwidth]{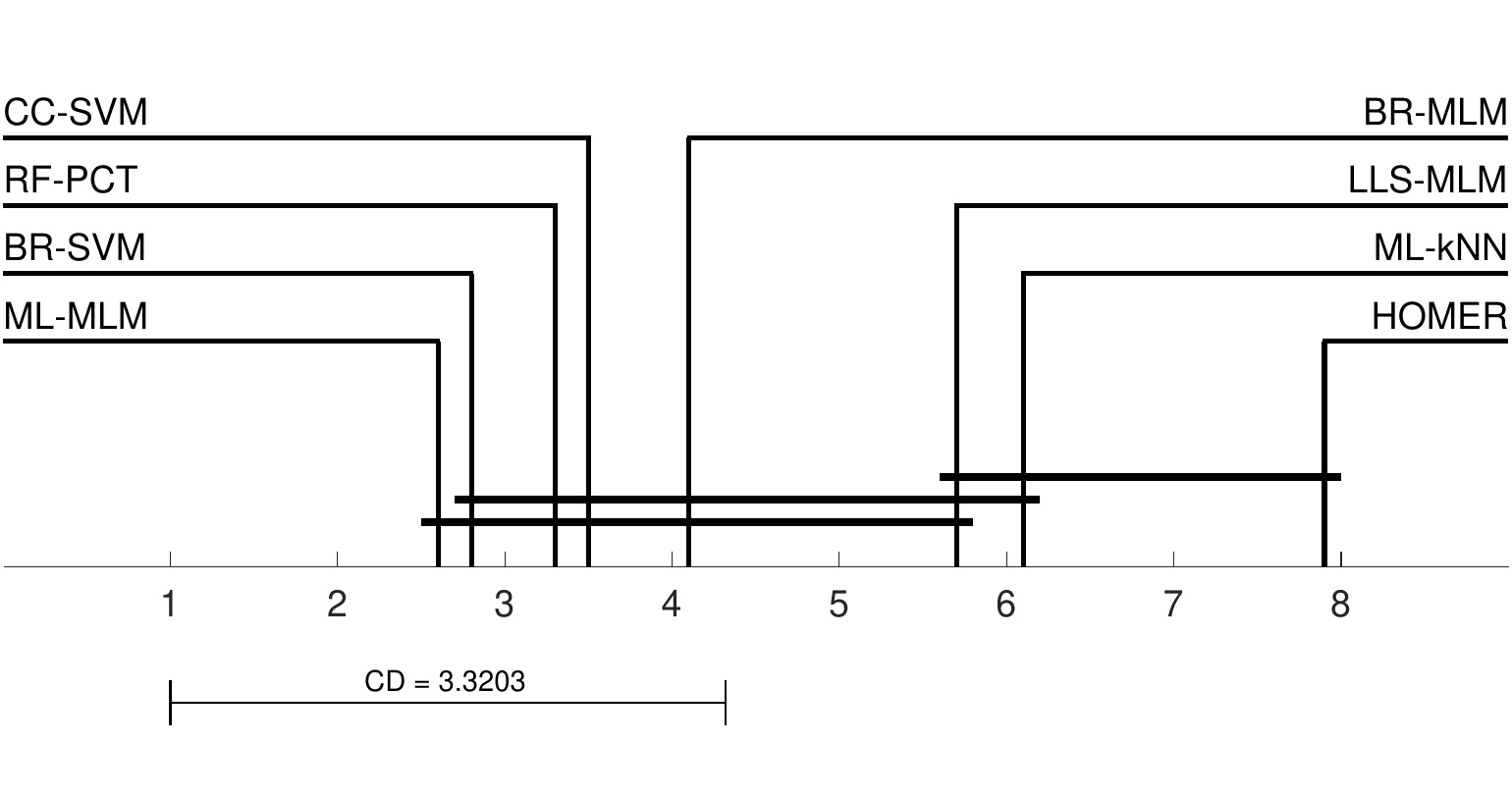}
    \caption{\textsc{Ranking Loss}}
    \label{fig:ranking_loss}
  \end{subfigure}
  ~~~~~
  \begin{subfigure}{0.45\textwidth}
    \includegraphics[width=\textwidth]{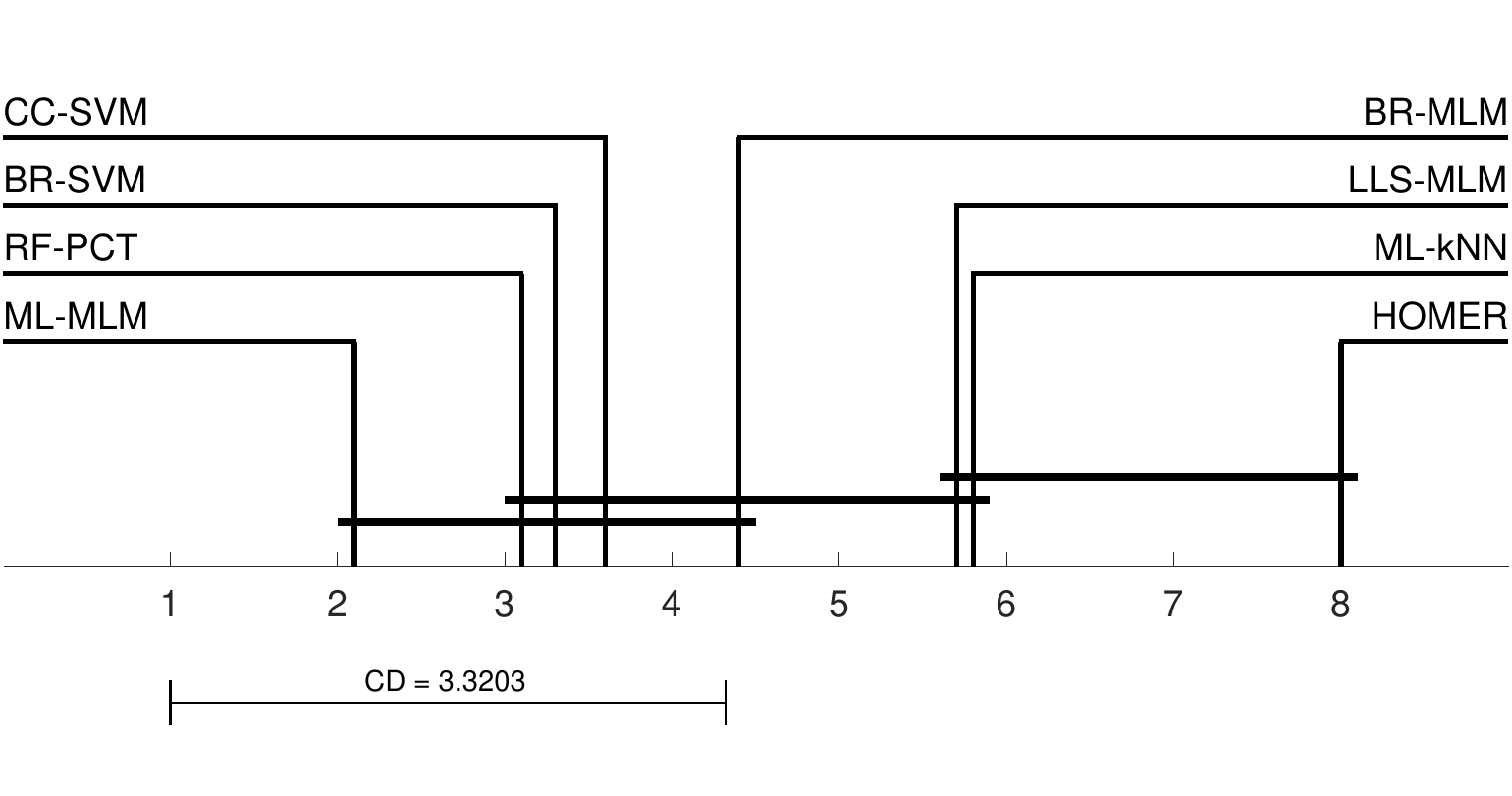}
        \caption{\textsc{Coverage}}
    \label{fig:coverage}
  \end{subfigure}

  \begin{subfigure}{0.45\textwidth}
    \includegraphics[width=\textwidth]{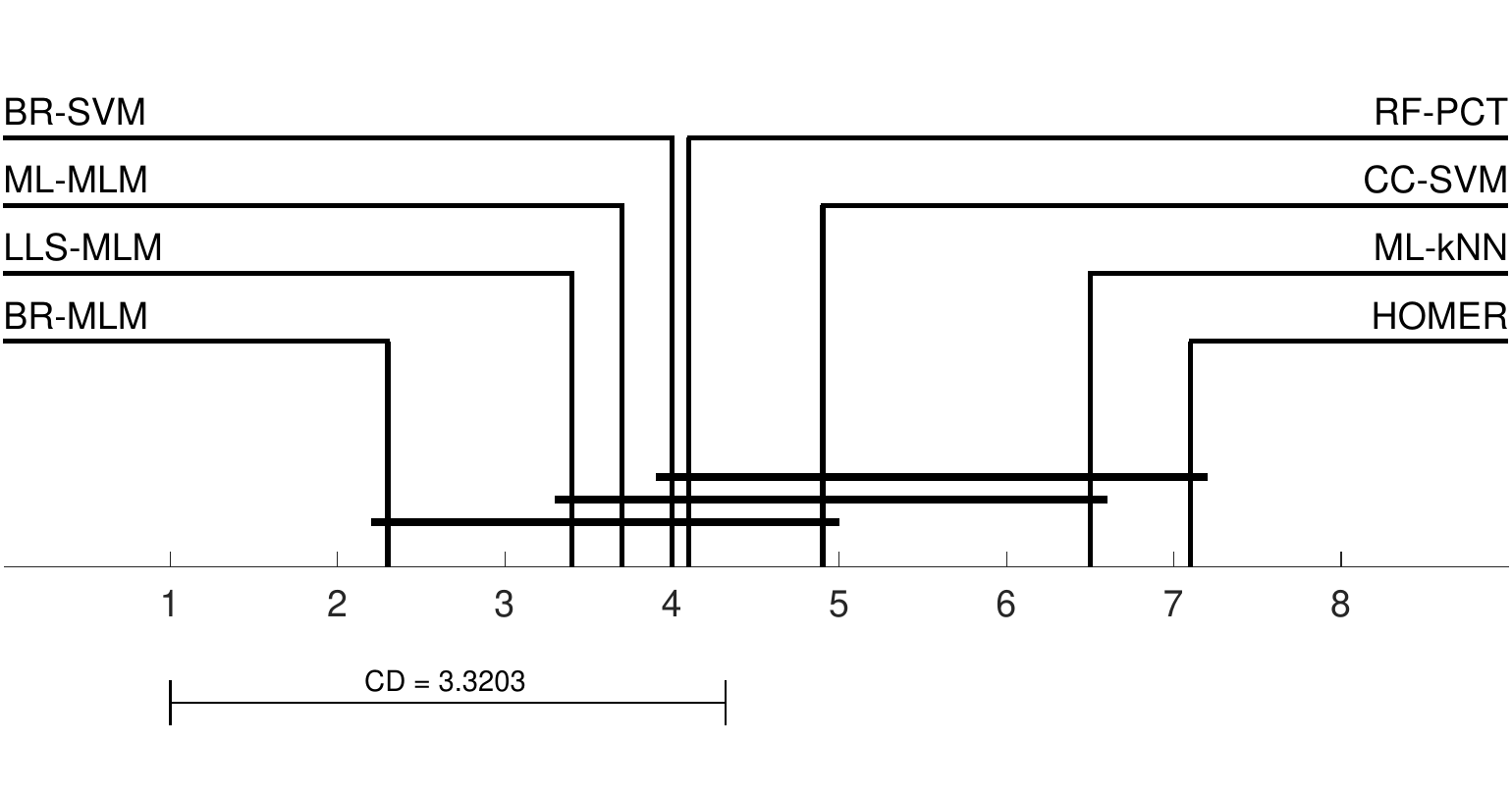}
    \caption{\textsc{One Error}}
    \label{fig:ranking_loss}
  \end{subfigure}
  ~~~~~
  \begin{subfigure}{0.45\textwidth}
    \includegraphics[width=\textwidth]{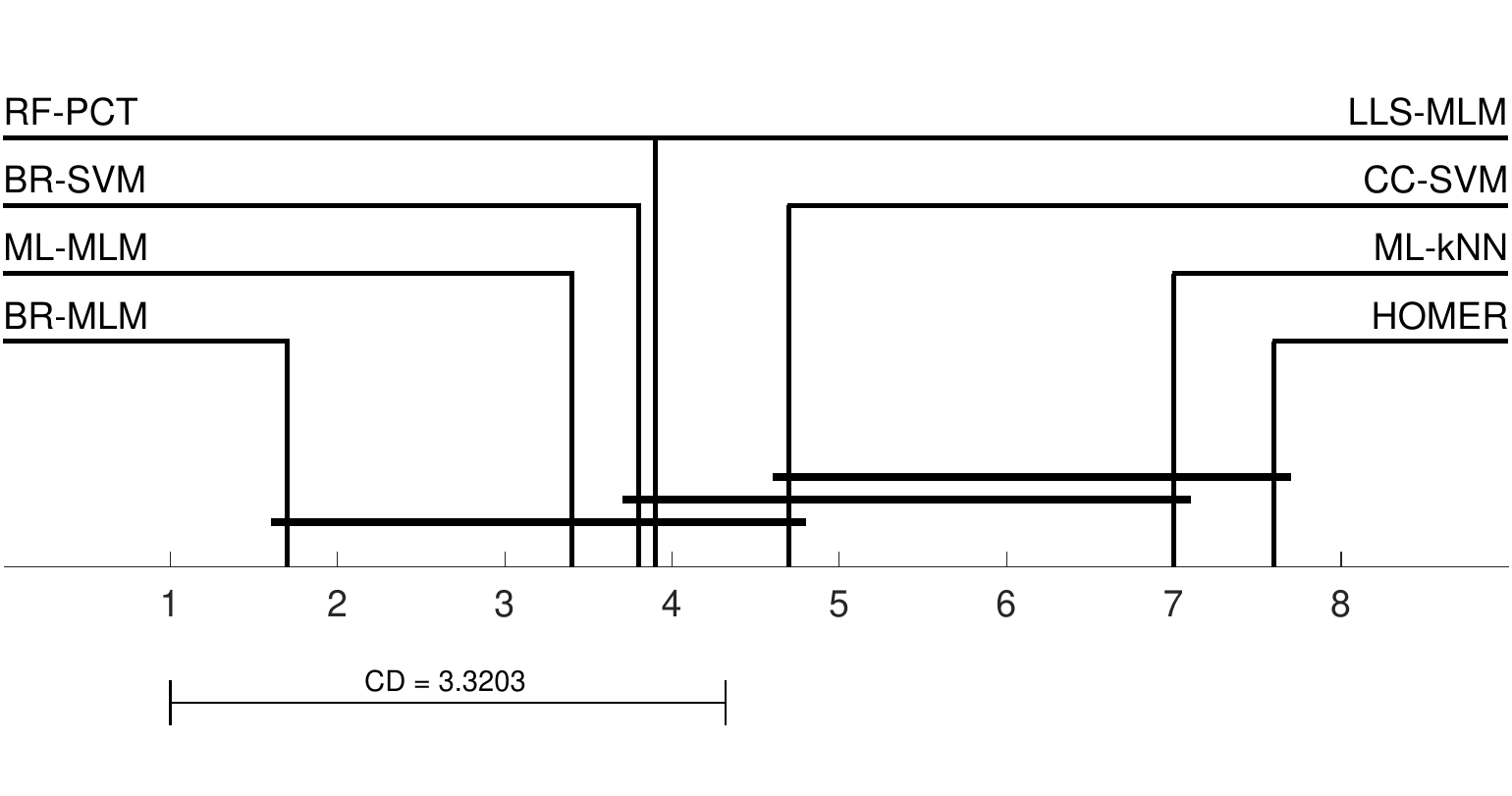}
        \caption{\textsc{Average Precision}}
    \label{fig:coverage}
  \end{subfigure}
  
	    \caption{Results for ranking-based metrics.} %\textcolor{red}{NOTE: iDMC = \mname, I will update the figure when we have consensus of the method name.}}
	\label{fig:ranking}
\end{figure}
 
% Overview
For all four ranking-based metrics, the Friedman null hypothesis was rejected with a clear margin. Therefore, we performed the Nemenyi post hoc tests for all four result tables (Tables \ref{tab:rl}-\ref{tab:ap} in Appendix \ref{app:ranking}). Statistical test results for these tests are presented in Figure \ref{fig:ranking} with the Nemenyi diagram, where the horizontal axis represents the average rank of a method for a given metric. A lower rank refers to better performance. A horizontal connection line on top of the axis indicates which methods do not have statistically significant differences according to the Nemenyi post hoc test.

% Ranking results figure
In Figures \ref{fig:ranking_loss} and \ref{fig:coverage}, we can see that ML-MLM have the highest average ranks for the \textsc{Ranking Loss} and \textsc{Coverage} metrics and the differences to ML-kNN and HOMER are statistically significant. In addition, there is a statistically significant difference between ML-MLM and LLS-MLM with respect to \textsc{Coverage}. For \textsc{One Error} and \textsc{Average Precision}, BR-MLM has the highest rank. For these metrics, BR-MLM and ML-MLM are both in the first group, thus having no statistically significant differences. Also, for these two ranking-based metrics, ML-MLM differs statistically from the ML-kNN and HOMER.

% Observations from the ranking results and dataset characteristics
From Tables \ref{tab:rl} - \ref{tab:ap} and the data characteristics Table \ref{tab:data}, a few observations can be made. All DRB methods have nearly perfect scores for all ranking-based metrics for the Tmc2007 dataset. In Table \ref{tab:data}, this dataset has $L_n = 0$, which means that all the label vectors in the test set were presented in the training set. Furthermore, consider the top four datasets with the lowest ratio $L_n/N_{ts}$: Tmc2007 (0), Scene (0.001), Emotions (0.020), and Yeast (0.041). For these datasets, ML-MLM outperforms the state-of-the-art RF-PCT method for all ranking-based metrics. This trend can also be observed at the upper end of the $L_n/N_{ts}$ ratios. For example, RF-PCT has better ranking performance than ML-MLM for the Delicious (0.972) and Enron (0.389) datasets.

% Pros and Cons of ML-MLM according on ranking results
According to the results, ML-MLM's performance is the strongest for \textsc{Coverage}. Based on solely the average ranks, an average rank difference between ML-MLM and RF-PCT is largest for \textsc{Coverage}. Good performance in respect of this metric indicates that if thresholding for ML-MLM is properly set, it should give an edge to ML-MLM in the bipartition-based comparison at least for the \textsc{Accuracy} metric.

% Summary, focus on ML-MLM and RF-PCT (state-of-the-art)
In summary, for the ranking-based metrics, there are no statistically significant differences between ML-MLM and RF-PCT. However, ML-MLM has a better average rank than RF-PCT for each of the metrics. 

\subsubsection{Bipartition-based metrics}

\begin{figure}[t]
  %\centering
  \begin{subfigure}{0.45\textwidth}
    \includegraphics[width=\textwidth]{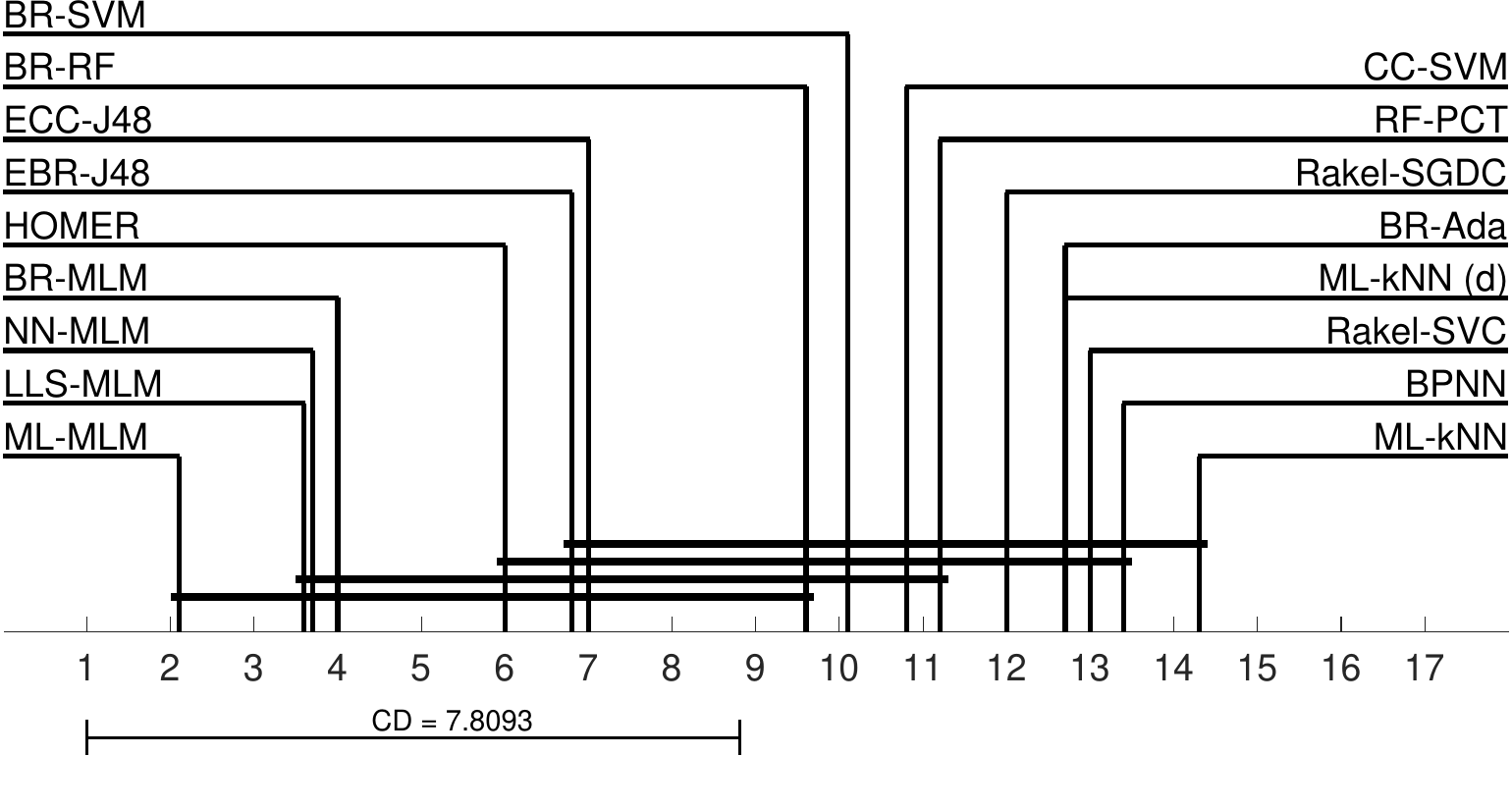}
    \caption{\textsc{Accuracy}}
    \label{fig:acc}
  \end{subfigure}
  ~~~~~
  \begin{subfigure}{0.45\textwidth}
    \includegraphics[width=\textwidth]{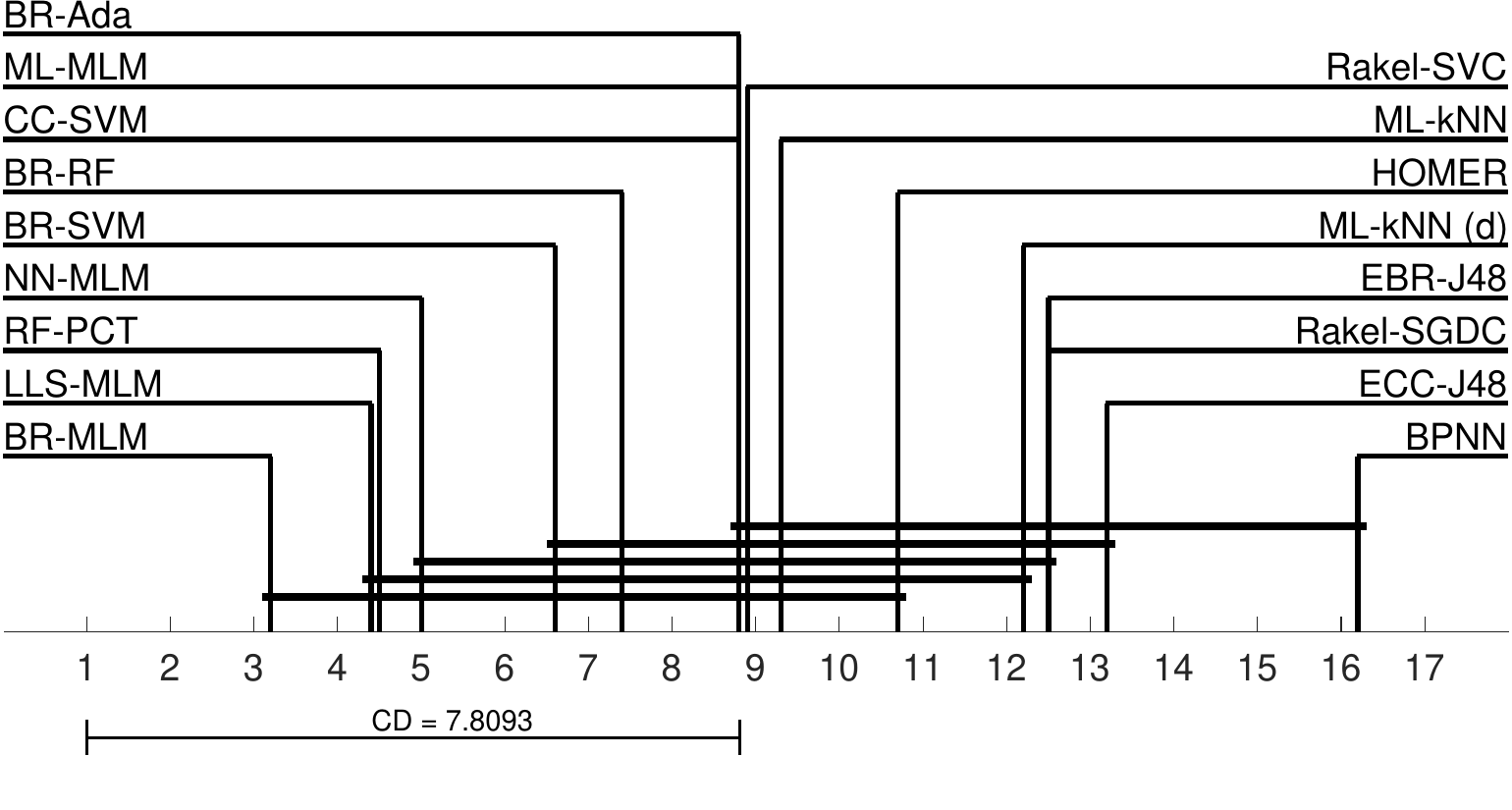}
        \caption{\textsc{Hamming Loss}}
    \label{fig:hl}
  \end{subfigure}

  \begin{subfigure}{0.45\textwidth}
    \includegraphics[width=\textwidth]{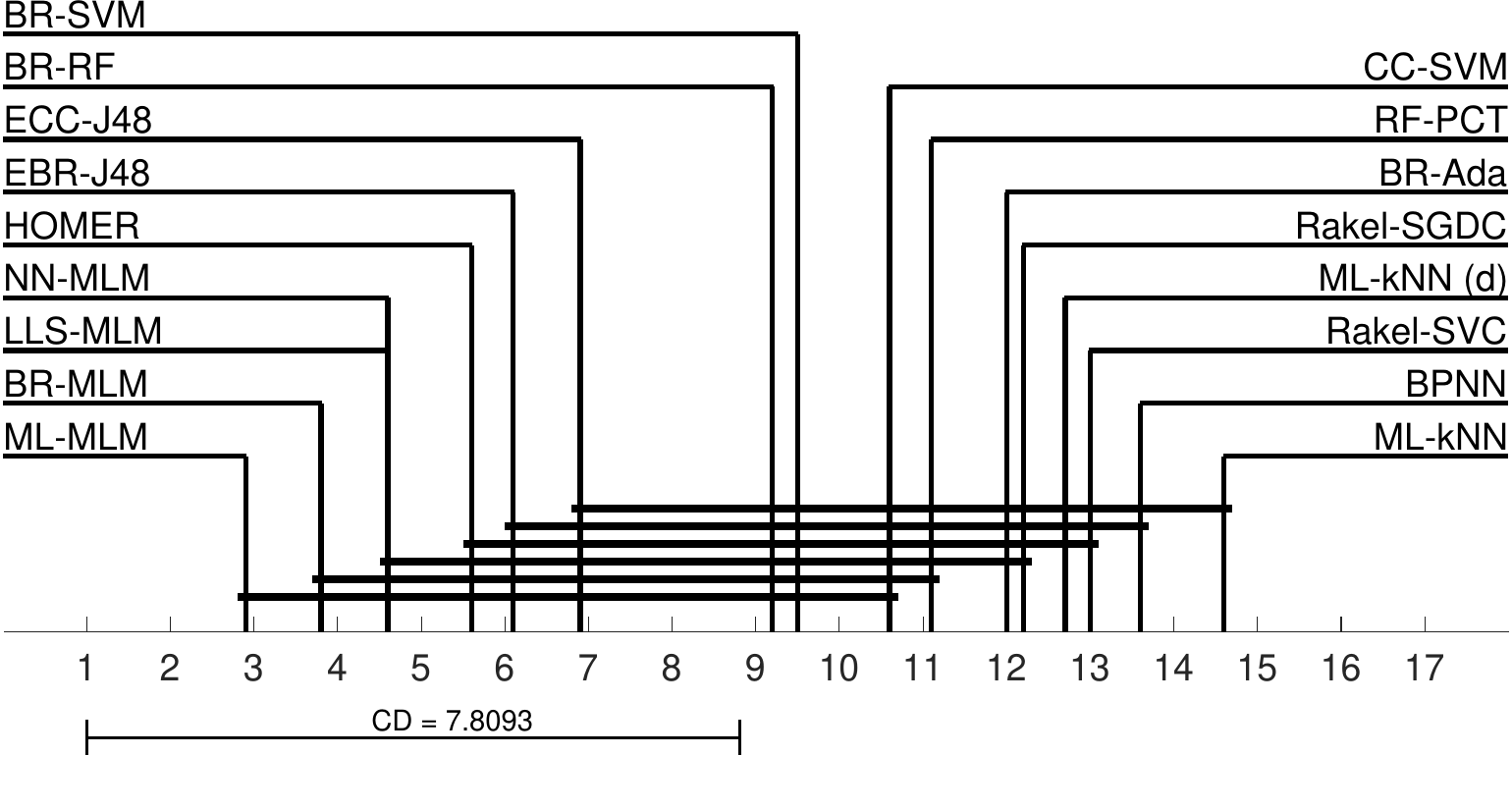}
    \caption{\textsc{Micro F1}}
    \label{fig:microf1}
  \end{subfigure}
  ~~~~~
  \begin{subfigure}{0.45\textwidth}
    \includegraphics[width=\textwidth]{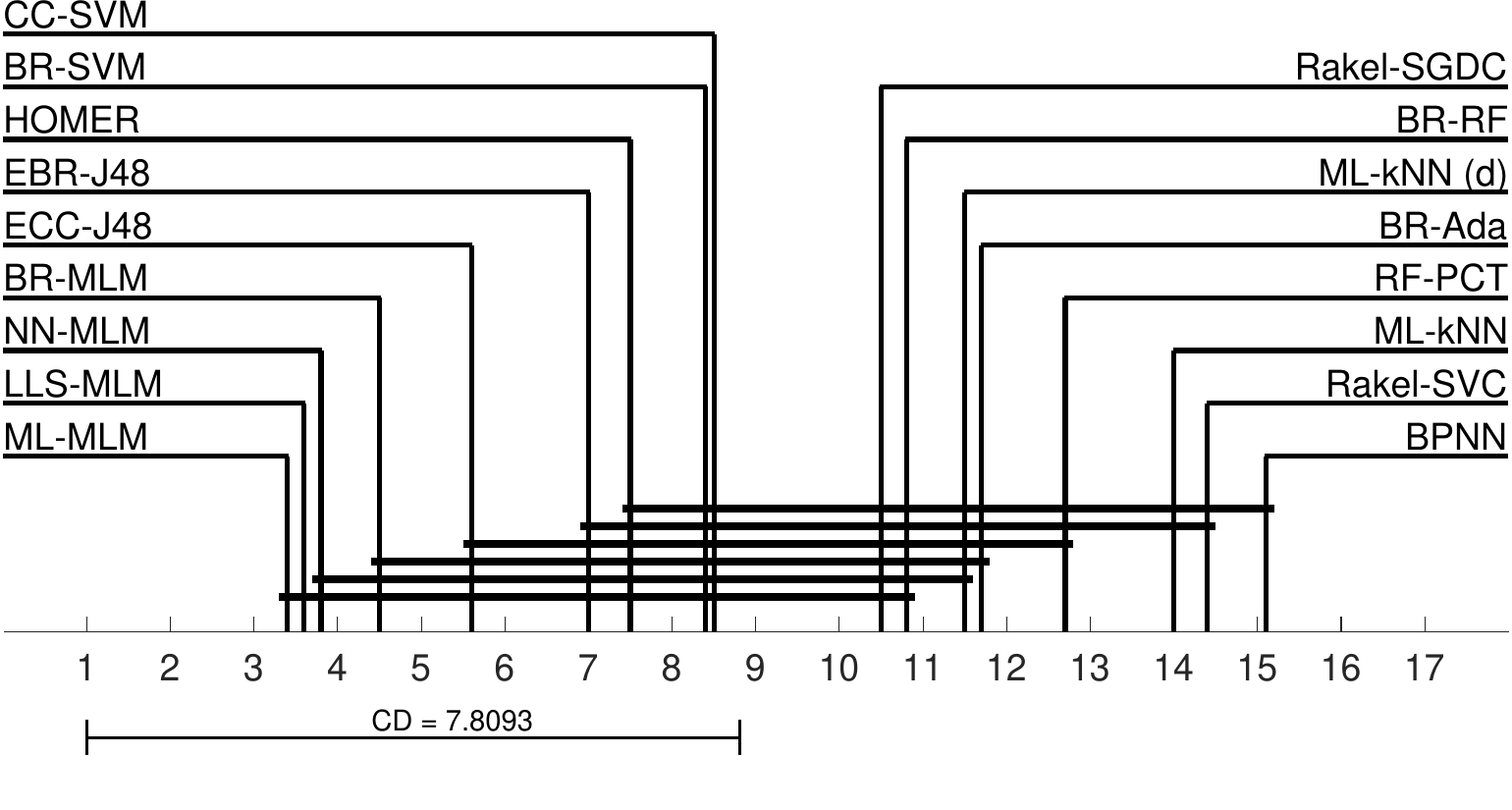}
        \caption{\textsc{Macro F1}}
    \label{fig:macrof1}
  \end{subfigure}
  
	    \caption{Results for bipartition-based metrics.} %\textcolor{red}{NOTE: iDMC = \mname, I will update the figure when we have consensus of the method name.}}
	\label{fig:bipartition}
\end{figure}

% Overview
The Friedman null hypothesis tests were also rejected with clear margins for all the bipartition-based results. Hence, similar to Section \ref{sec:ranking}, the Nemenyi post hoc test procedure was performed for Tables \ref{tab:acc}-\ref{tab:macrof1}. The results of these tests are again summarized with the critical diagram visualizations given in Figure \ref{fig:bipartition}. Note that differently from Section \ref{sec:ranking}, we have now included NN-MLM, {BR-J48, CC-J48, BR-Ada, BR-RF, Rakel-SCV, Rakel-SGDC, ML-kNN (d), ECC-J48, EBR-J48 and BP-NN in this comparison, so we have 17 methods in total, compared to eight in the previous one. The critical distance is therefore larger in statistical testing with respect to this section.}

% Bipartition results figure
According to Figure \ref{fig:bipartition}, ML-MLM has the best average rank for the metrics \textsc{Accuracy}, \textsc{Micro F1}, and \textsc{Macro F1}. Moreover, a ranking difference between ML-MLM and RF-PCT is statistically significant for these three metrics. For \textsc{Hamming Loss}, BR-MLM has the best performance and is statistically significant with respect to {ML-kNN(d), EBR-J48, Rakel-SGDC, ECC-J48, and BPNN}. Although ML-MLM has a worse average rank for \textsc{Hamming Loss} than BR-MLM and RF-PCT, this difference is not statistically significant. Note that there are no statistically significant differences between the DRB methods. {The ensemble-based construction of the CC (ECC-J48) has a worse rank than the vanilla CC. Moreover, the ensemble-based methods gropu, ECC-J48, EBR-J48, and HOMER, have slightly worse ranks than the DBR-based methods for the metrics \textsc{Accuracy}, \textsc{Micro F1}, and \textsc{Macro F1}. Note that the ensemble methods are statistically in the same group as the DBR-based methods with these metrics. ML-kNN and BPNN are the methods that perform the worst in the bipartition-based comparison. Consequently, the off-the-shelf implementation of BPNN is not well suited for small- to moderate-scale datasets. BPNN requires, as neural networks in general, enough data and careful fine-tuning of the parameters.}  % Moreover, ML-MLM and HOMER do not differ statistically in any of the metrics.

% Pros of ML-MLM based on bipartition results
{As expected, based on the high \textsc{Coverage}-performance, ML-MLM performs best in \textsc{Accuracy} compared to the other methods. A statistical interpretation between ML-MLM and a group of methods ECC-J48, EBR-J48, and HOMER is the same for \textsc{Accuracy}, \textsc{Macro}, and \textsc{Micro}. However, if we solely inspect the average rank differences between this group and ML-MLM, regarding \textsc{Accuracy} these differences are larger. Moreover, from Table \ref{tab:acc} and Table \ref{tab:microf1}, we can see that ML-MLM have the highest \textsc{Accuracy} value for five out of ten datasets and the highest \textsc{Micro F1} value for six out of ten datasets. Moreover, the results for the label-wise bipartition-based metrics, \textsc{Micro F1} and \textsc{Macro F1}, show that the predicted set of labels outputted by ML-MLM is accurate and well balanced for all labels.
}
%As expected, based on the high \textsc{Coverage}-performance, \mname performs best in \textsc{Accuracy} compared to the other methods. A statistical interpretation between \mname and a group of methods RF-PCT, CC, ML-kNN, and BR, is the same for \textsc{Accuracy} and \textsc{Micro}. However, if we solely inspect the average rank differences between this group and \mname, regarding \textsc{Accuracy} these differences are larger. Moreover, from Figure \ref{fig:acc} and Figure \ref{fig:microf1}, we can see that \mname have the highest \textsc{Accuracy} value for six out of ten datasets and the highest \textsc{Micro F1} value for seven out of ten datasets. The results for label-wise bipartition-based metrics, \textsc{Micro F1} and \textsc{Macro F1}, show that the predicted set of labels outputted by \mname is accurate and well-balanced for all labels. 

% Cons of ML-MLM based on bipartition results

{A minor drawback of ML-MLM is the performance compared to \textsc{Hamming Loss}, for which the average rank is worse compared to all the other DRB methods, BR-RF, BR-SVM, and RF-PCT. Since ML-MLM performs better in \textsc{Accuracy} and worse in \textsc{Hamming Loss}, it could be greedier than the other methods when choosing the labels, increasing the number of false positives. However, since the average rank is not statistically different from the higher performing methods and is better compared to Rakel-SVC, ML-kNN, HOMER, ML-kNN(d), EBR-J48, Rakel-SGDC, ECC-J48, and BPNN, we argue that this trade-off is reasonable. Note that even the \textsc{Hamming Loss}'s absolute values are small for all the experimented methods in many of the datasets. This is mostly due to the normalization with the number of all possible labels. 
}

% Observations from the bipartition results and dataset characteristics
%Similarly to previous result section, a few observations for the results and data characteristics can be made based for Tables \ref{tab:acc} - \ref{tab:macrof1} and Table \ref{tab:data}. 

% Summary
{In general, the results show that ML-MLM is statistically better than the state-of-the-art RF-PCT method for three out of four bipartition-based metrics and it is statistically equal for the \textsc{Hamming Loss}. Moreover, ML-MLM has a better rank than the ensemble methods (ECC-J48, EBR-J48 and HOMER), for all bipartition-based metrics, but this difference was not identified as statistically significant. All DRB methods perform well in the comparison, and also the simplest method (NN-MLM), which is statistically equal in three metrics and better in one metric than RF-PCT.
}

\subsection{Assessing uncertainty and interpreting ML-MLM} \label{sec:interpretation}

% Overview
Understanding how ML-MLM predicts can be interpreted via the values of the selected power parameters and the predicted distances. These results are summarized in Table \ref{tab:P}, where the distance corresponding to the predicted nearest neighbor is squared and averaged over the test set for each dataset. We squared the predicted distance in Table \ref{tab:P} for better interpretability of the label differences (see Appendix \ref{app:demo} for more details). In Figure \ref{fig:pred_dists}, the distributions of the predicted distances (without squaring) are presented with box plots.

% Interpretation with the selected power parameter
From Table \ref{tab:P}, we can see that the selected power parameter values vary significantly between the datasets. The smallest value was set for the Scene dataset ($P = 4$), while the largest value was set for the Medical dataset ($P \approx 222.9$). In the latter, ML-MLM is predicting the set of labels highly similar to NN-MLM, since almost all the weight is given to the predicted nearest neighbor. We verified this for Medical by comparing bipartition-based metric results between NN-MLM and ML-MLM with the local RCut thresholding. The results for this comparison were identical. The results regarding Medical and the ML-MLM and NN-MLM methods in Tables \ref{tab:acc}-\ref{tab:macrof1} are different due to ML-MLM's using the cardinality-based thresholding (Eq. \ref{eq:card_t}) instead of the local RCut thresholding. For the Medical dataset, NN-MLM gives better results for three out of four metrics, which indicates that the cardinality-based thresholding could be suboptimal for large power parameter values. 

% Interpretation with the selected power parameter + predicted distances
From Table \ref{tab:P} and the characteristics of the dataset (Table \ref{tab:data}), we can notice that ML-MLM may select large power parameter values for highly different types of datasets, such as Bibtex ($P = 128$) and Tmc2007 ($P = 59.7$). However, if we combine information from the predicted distances and the power parameter values, we can interpret for Tmc2007 that ML-MLM does not benefit much from increasing the contribution of other reference points in the convex combination, since the nearest predicted neighbor already contains a significant amount of information for a nearly optimal prediction. Since both the predicted distances and the power parameter are large for Bibtex, ML-MLM is mostly extrapolating and does not benefit much from combining label information from multiple reference points.

% Uncertainty of MLC prediction
Regarding the predicted distances in Table \ref{tab:P} and the overall ranking results (Tables \ref{tab:rl} - \ref{tab:ap}) and classification results (Tables \ref{tab:acc} - \ref{tab:macrof1}) of ML-MLM, we can summarize the following. For a given instance, the predicted distance can be used directly as an uncertainty estimate of how reliable the ranking or classification is. Higher distances reflect that the instance is probably far away from the data and the model is extrapolating. Thus, the prediction is probably unreliable. In this way, the predictions could be straightforwardly categorized in different uncertainty categories (e.g., low: $distance < 1$, medium: $1 < distance < \sqrt{2}$, high: $distance >\sqrt{2}$) based on the predicted distance(s). Moreover, for a given dataset, we could use the squared predicted distance to characterize the difficulty of the ML problem in an intuitive way.

\begin{table}[h]
\caption{Selected power parameter $P$ values and average minimum predicted squared Euclidean distances for ML-MLM$\!\!$.}\label{tab:P}
\normalsize
\centering
\bgroup
\def\arraystretch{1.2}%
\begin{adjustbox}{max width=1\columnwidth}
\begin{tabular}{l|rrrrrrrrrrrrr}
Dataset & Medical & Emotions & Enron & Scene & Yeast & Corel5k & Bibtex & Delicious & Tmc2007 & Mediamill\\ \hline
P & $2^{7.8}$ & $2^{2.9}$ & $2^{3.8}$ & $2^{2}$ & $2^{3.2}$ & $2^{4.1}$ & $2^{7}$ & $2^{4.2}$ & $2^{5.9}$ & $2^{4.4}$ \\
 & $\approx 222.9$ & $\approx 7.5$ & $\approx 13.9$ & $\approx 4$ & $\approx 9.2$ & $\approx 17.1$ & $\approx 128$ & $\approx 18.4$ & $\approx 59.7$ & $\approx 21.1$ \\
Distance &  $0.6 \pm 0.4$ &  $0.8 \pm 0.4$ &  $1.7 \pm 1.1$ &  $0.3 \pm 0.3$ &  $2.1 \pm 0.9$ &  $3.7 \pm 0.5$ &  $1.9 \pm 0.9$ &  $16.6 \pm 5.4$ &  $0.2 \pm 0.2$ &  $2.4 \pm 1.1$ \\ \hline
\end{tabular}
\end{adjustbox}
\egroup
\end{table}

\begin{figure}[h]
  \centering
    \includegraphics[width=\textwidth]{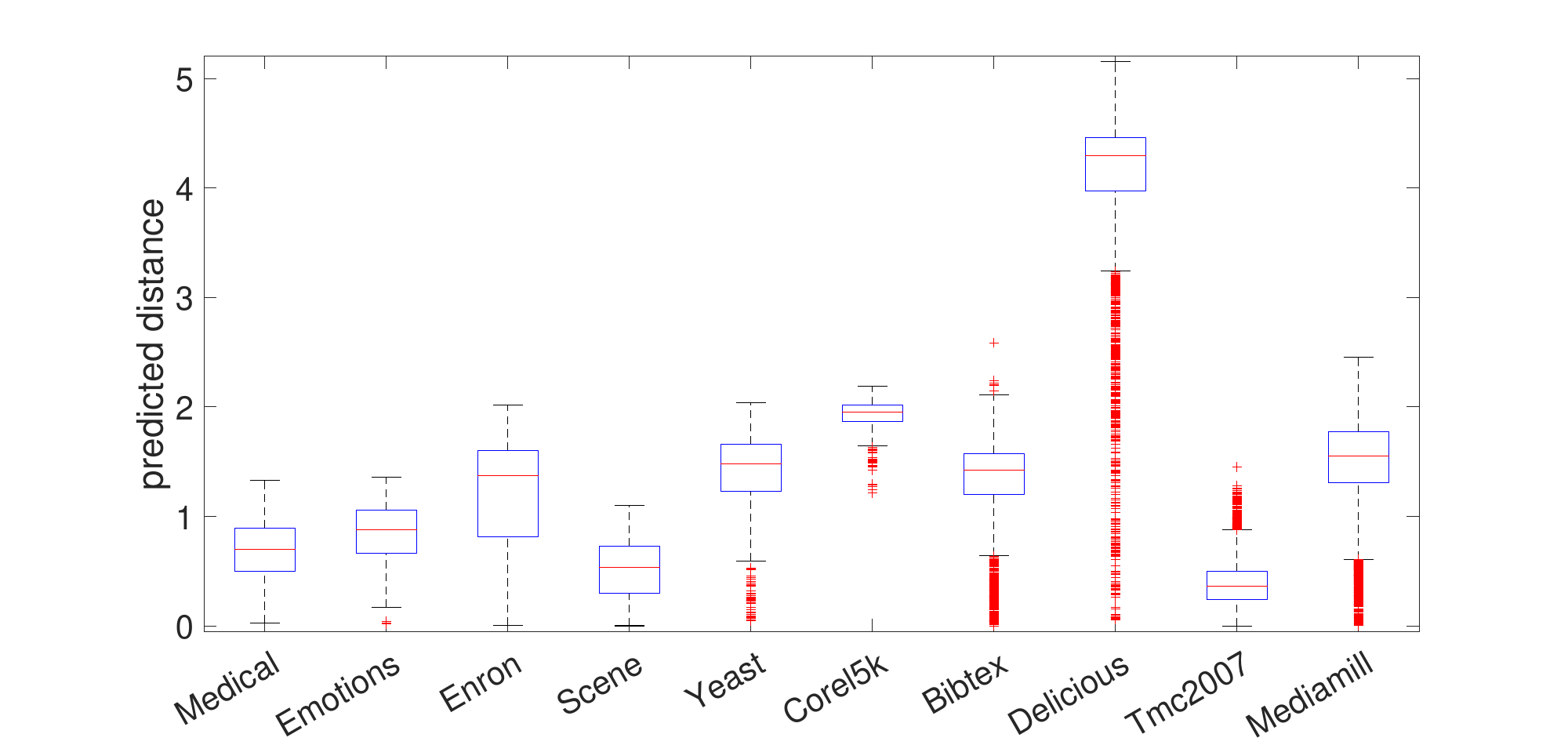}
    \caption{Distance regression prediction's minimum distances for the test sets. Predicted distances closer to zero reflect that the distance regression model is interpolating more. Distance regression model's overall interpolation ability is strongest for Tmc2007 and worst for Delicious.}
	\label{fig:pred_dists}
\end{figure}

% How selected power parameter values characterize data?

\subsection{Time complexities of distance regression based methods}\label{sec:cost}

% Distance regression training
Distance regression step with Moore-Penrose pseudoinverse has a $\mathcal{O}(NK^2)$ time complexity. Since we select nearly all distinct points as reference points, the time complexity for LLS-MLM, ML-MLM, and NN-MLM is $\mathcal{O}(N^3)$ for datasets that have $N$ distinct input instances. With these same assumptions, training time complexity for BR-MLM is $\mathcal{O}(LN^3)$, since a distinct Moore-Penrose pseudoinverse is needed for each label. Therefore, for the DBR methods, the BR-MLM method has the highest time complexity in training due to the $L$ dependency. From now on, we will handle the prediction time complexity analysis with the assumption of distinct input instances, meaning that $K = N$. This assumption corresponds to the worst-case scenario regarding the time complexity. 

% LLS prediction
The LLS's linear system can be solved via an OLS solution, which has a cost of $\mathcal{O}(L^2N)$ \cite{hamalainen2020minimal}. Therefore, for a large number of labels, this cost dominates the overall time complexity, and it has to be considered in MLC problems. Similar to ML-MLM, LLS predicts label space distances which has a cost of $\mathcal{O}(N^2)$. Then, the computational cost for LLS prediction is $\mathcal{O}(L^2N + N^2)$. Thus, in the MLC context, ML-MLM has a lighter computational cost than LLS with respect to the number of labels.

% NN-MLM prediction
For the DBR methods, NN-MLM is the fastest in prediction, because, after the given distance predictions, the predicted set labels can be searched with $\mathcal{O}(L_u)$ cost \cite{Mesquita2017}, where $L_u$ is the number of distinct label vectors in the training set. However, due to the distance prediction step, the overall computational complexity is still $\mathcal{O}(N^2)$ for the NN-MLM.

% ML-MLM prediction
Similar to NN-MLM, ML-MLM's prediction time complexity is driven by the distance prediction step. Therefore, ML-MLM's time complexity is also $\mathcal{O}(N^2)$. A naive implementation of step 2 (Algorithm \ref{alg:idwmlmpred}) requires $\mathcal{O}(KN)$ cost. However, similar to NN-MLM, this can be reduced to $\mathcal{O}(C_u L_u)$, where $L_u$ is the number of unique label vectors in the training set and $C_u$ is the label cardinality of these label vectors. This label cardinality dependency is due to applying sparse structures for label vectors. Furthermore, for the MLC problems where label cardinality is small, this complexity is reduced to $\mathcal{O}(L_u)$.

In addition to training, the BR-MLM method also has the highest computational cost for prediction. The time complexity for BR-MLM's prediction is $\mathcal{O}(L^2N)$ because distance predictions are estimated for each label separately with $L$ models.

As a reminder, note that these training and prediction time complexities can be significantly alleviated with the clustering-based distance regression construction \cite{hamalainen2021instance}. Here, we were interested in comparing the core models' costs under simplifying assumptions.

\section{Discussion}\label{sec:discussion}

% Ranking results discussion

% Bipartition results discussion

% Thresholding choices
All DRB methods performed well in the bipartition-based metrics comparison. This result indicates that the selected thresholding approaches are well suited for the DRB methods. The proposed thresholding method, local RCut, gives a straightforward way to set the threshold adaptively during prediction via the cardinality of the NN-MLM's prediction. Note that this does not add complexity to the prediction if local RCut thresholds a DRB model with the same core distance regression model. We observed that the local RCut thresholding outperforms the standard fixed $0.5$ thresholding for the DRB methods. Therefore, we recommend avoiding this fixed thresholding for the DRB methods. This observation aligns with the results and recommendations presented in \cite{alotaibi2021multi}. To improve ML-MLM classification performance for higher power parameter values, a hybrid approach of the global and local RCut thresholding could be considered.

% Differences and benefits of the DRB methods
The proposed method ML-MLM performed overall better than the other DRB methods in the comparison based on the average ranks. However, ML-MLM was statistically different from other DRB methods only regarding the \textsc{Coverage} metric. In Section \ref{sec:background}, we showed that the NN-MLM method's prediction does not guarantee multilateration problem solving for MLC. However, based on highly similar classification results (Figure \ref{fig:bipartition} and Tables \ref{tab:acc}-\ref{tab:macrof1}) for the LLS-MLM and NN-MLM, this simple nearest neighbor based heuristic is very close to solving the multilateration problem similarly as the more complex LLS approach. Based on plainly on these results, there seems to be no reason to use LLS-MLM instead of NN-MLM. However, the main limitation of NN-MLM is that it is not able to provide the ranking/scoring of the predicted labels, and therefore its usability is limited. In addition, NN-MLM's predictions are limited to the training set label vectors. Moreover, although the DRB methods predict the set of labels via highly different approaches, they all performed well in the comparison. This indicates that the DRB construction is beneficial for the MLC problems and that the distance predictions from this can be used in many ways to end up with a good solution.

% What is the value of using ML-MLM instead of other DRB methods?
Taking into account the ranking and classification performances and training/prediction time complexities, the most viable option of the DRB methods is ML-MLM. As noted in the previous paragraph, there are clear benefits in using ML-MLM instead of NN-MLM and BR-MLM. When comparing ML-MLM to LLS-MLM, ML-MLM improves the ranking performance for the metrics \textsc{Ranking Loss} and \textsc{Coverate}. Moreover, the prediction phase time complexity of ML-MLM is linear with respect to the number of labels compared to the quadratic time complexity of LLS-MLM.

% What is the value of using ML-MLM over RF-PCT?

% Challenges of ML-MLM

\section{Conclusion}\label{sec:conclusion}

% Overview of the main contribution (ML-MLM + its results compared to RF-PCT)
In this paper, we proposed a simple distance regression-based method for multi-label learning referred to as ML-MLM. The core idea of the proposed method is to learn label ranking via distance regression and inverse distance weighting. Besides being one of the simplest methods in the multi-label classification (MLC) literature, our experimental evaluation showed that it achieves the state-of-the-art level of multi-label ranking and classification performance for small to moderate-sized MLC problems. Furthermore, in the classification performance evaluation, the proposed method outperformed a random forest-based method, which is considered one of the state-of-the-art methods in the latest MLC literature. In the ranking performance evaluation, the proposed method performed equally well as the random forest-based method. Moreover, we showed that the proposed method outperforms the ML-kNN method, which is considered to belong to the same group as our method in the MLC taxonomy.

% Justifying the cubic complexity in this paper
% Tommi: Added papers which include more efficient solvers for that stage.
Although the proposed method's training time complexity is independent of the number of labels, it has a cubic training time complexity with respect to the number of observations for the basic Cholesky decomposition methods. %In the current formalism, 
With the basic algorithms, it is a bottleneck of the proposed method concerning its scalability. However, our main goal in this work was to show a proof of concept that a simple distance regression-based approach can achieve high accuracy in MLC problems. Therefore, we omitted reducing the cubic training time complexity that arises from solving a large-scale ordinary least squares problem. Moreover, even though the training time is cubic, sophisticated implementations exist for these problems because of their popularity and long history. For example, random approximation and GPU computing-oriented methods \cite{linja2020randomized,lai2019maximally} can be used to improve the efficiency of this step. 

% Conclusion about distance regression's potential for multi-output learning
In general, the results showed and strengthened previous observations that exploiting the linear distance mapping between feature and multidimensional output spaces is beneficial for multi-output learning problems. In this paper's formalism, connections between all the distinct points are represented via these distance mappings. This means that we simplify the learning on a feature level with dissimilarity but explore the whole dissimilarity structure of the data for prediction. In other words, we assume that features are equally important, but holding on the possibility that every dissimilarity between the observations might contribute to the prediction.

%Check place of the paragraph below
Clearly, the proposed method could be improved using similar technical enlargements and modifications, as summarized in Section \ref{sec:relatedwork}. Especially, the ensemble techniques already depicted in \cite{hamalainen2021instance} with the MLM techniques can be used to introduce more specific models for subsets of the training data that could be selected based on features (input) or labels (output), or their dependencies (see Section \ref{sec:relatedwork}). For both the global and local models, feature selection can be used to further specify the models to encapsulate the relevant behavior of the multi-label data. Moreover, because the core construct of MLM is distance regression, which is based on pairwise dissimilarities in the input space, other data modalities and corresponding dissimilarity measures should be further tested. For textual data, which was considered for the multi-label classification tasks in \cite{liu2015multi,jiang2017multi,almeida2018emotion,parwez2019multi,du2019ml,zhang2020multi,chen2022multi}, one could apply many similarity measures \cite{wang2020measurement}. The same is true for imaging data, considered in \cite{chen2019deep,chen2020deep,pan2020multi,bi2020multi}, by using, e.g., a patch-based dissimilarity \cite{amelio2016patch}.
%Conclusions may be used to restate your hypothesis or research question, restate your major findings, explain the relevance and the added value of your work, highlight any limitations of your study, describe future directions for research and recommendations. 

%In some disciplines use of Discussion or 'Conclusion' is interchangeable. It is not mandatory to use both. Please refer to Journal-level guidance for any specific requirements. 

%\bmhead{Supplementary information}

%If your article has accompanying supplementary file/s please state so here. 

%Authors reporting data from electrophoretic gels and blots should supply the full unprocessed scans for key as part of their Supplementary information. This may be requested by the editorial team/s if it is missing.

%Please refer to Journal-level guidance for any specific requirements.

%\bmhead{Acknowledgments}

%Acknowledgments are not compulsory. Where included they should be brief. Grant or contribution numbers may be acknowledged.

%Please refer to Journal-level guidance for any specific requirements.

\section*{Author Contributions} JH developed the proposed methods, wrote the implementations, designed and performed the experiments, analyzed the results, and wrote the majority of the paper. {AH performed and designed the experiments regarding the MEKA wrapper implementations, contributed to the writing of the experimental section.} AHS conceptualized the theoretical results, wrote the proofs, and contributed to the writing of the method sections and the initial draft. TK performed literature search, wrote majority of the related work section, and contributed to writing of the initial draft. CLCM, TK and JPPG revised and commented on the initial draft. JPPG provided a reference implementation of the LLS algorithm. All authors contributed to the writing and editing of the final draft.

\section*{Funding} This work was supported by the Academy of Finland through the grant 351579 in the EuroHPC Research Programme.

\section*{Declarations}

The authors declare that they have no conflict of interest.

\section*{Data Availability}

The datasets that were used are publicly available.

\section*{Code Availability}

All codes are available from \textsl{Gitbub} \url{https://github.com/jookriha/ml-mlm}.%\ldots

%Some journals require declarations to be submitted in a standardised format. Please check the Instructions for Authors of the journal to which you are submitting to see if you need to complete this section. If yes, your manuscript must contain the following sections under the heading `Declarations':


%\bibliography{refs_esann2021,refs_new,related_work,Bibliography}
\bibliographystyle{unsrt}
\bibliography{all_refs}



\begin{appendices}
%\section{}\label{app:metrics}

\section{Proofs}\label{ap:proof}

For clarity, we split the proof of Proposition 1 into two parts, corresponding to the numbered statements. 

\vspace{0.5cm}

\noindent \textbf{Part I: for arbitrary $[\hat{\delta}_k]_k,$ NN-MLM does not yield optimal solutions}
\begin{proof}
To show that for arbitrary $[\hat{\delta}_k]_k$ NN-MLM does not always yield optimal solutions, it suffices to provide a counterexample. Thus, consider $L=2$, $\mathbf{y}_1=[0, 0]^\top, \mathbf{y}_2=[0, 1]^\top, \mathbf{y}_3=[1, 0]^\top, \mathbf{y}_4=[1, 1]^\top$, and $\mathbf{t}_i=\mathbf{y}_i$ for all $i$. Also, let us define $\hat{\delta}_1=1, \hat{\delta}_2=10, \hat{\delta}_3=2, \hat{\delta}_4=2$. 

In this case, the prediction from NN-MLM is $\hat{\mathbf{y}}=\mathbf{y}_1=[0, 0]^\top$ with associated loss equal to 
\begin{align*}
J(\mathbf{y}_1) & = (- \hat{\delta}_1^2)^2 + (1 - \hat{\delta}_2^2)^2 + (1 - \hat{\delta}_3^2)^2 + (2 - \hat{\delta}_4^2)^2\\
& = 1 + 99^2 + 3^2 + 2^2 = 9815
\end{align*}

Now, consider the value $J(\mathbf{y}_3)$:
\begin{align*}
J(\mathbf{y}_3) & = (1 - \hat{\delta}_1^2)^2 + (2 - \hat{\delta}_2^2)^2 + (0 - \hat{\delta}_3^2)^2 + (1 - \hat{\delta}_4^2)^2\\
& = 0 + 98^2 + 4^2 + 3^2 = 9629
\end{align*}
Therefore, since $J(\mathbf{y}_3) < J(\mathbf{y}_1)$, $\hat{\mathbf{y}}=\mathbf{y}_1$ is not an optimal solution to the multilateration problem in Equation 1.
\end{proof}

%\begin{lemma}
%Let $S$ be a indexed multiset of non-negative real numbers $S=\{s_i\}_{i=1}^n$ such that $s_1 \leq s_2 \leq  \dots \leq s_n$. Let $\pi$ be a permutation of $n$ elements. Then,
%$$
%\sum_{i=1}^n s_i s_{\pi(i)} \leq \sum_{i=1}^n s_i^2.
%$$
%\end{lemma}
%\begin{proof}
%It suffices to show that an optimal solution $\pi^\star$ to $\arg\max_\pi \sum_{i=1}^n s_i s_{\pi(i)}$ consists of the identity permutation $\pi^\star(i) = i$ for $i=1, \dots, n$.

%Note that $\argmax_\pi  \sum_{i=1}^n s_i s_{\pi(i)} = \argmin_\pi \sum_{i=1}^n (s_i - s_{\pi(i)})^2$:
%\begin{align*}
%& \argmin_\pi \sum_{i} (s_i - s_{\pi(i)})^2 \\
%&  =\argmin_\pi \sum_i s_i^2 - \sum_i 2s_{\pi(i)}s_i + \sum_i s_{\pi(i)}^2 \\
%&  =\argmax_\pi \sum_{i=1}^n s_i s_{\pi(i)},
%\end{align*}
%since $\sum_i s_i^2$ and $\sum_i s_{\pi(i)}^2$ are constant for any permutation $\pi$. Finally, note that the identity permutation $\pi^\star(i)=i$ is a minimizer to $\sum_{i=1}^n (s_i - s_{\pi^\star(i)})^2 = 0$.
%\end{proof}

\noindent \textbf{Part II: If $\hat{\delta}_k = \|\mathbf{y}' - \mathbf{t}_k\|$ for some $\mathbf{y}' \in \mathbb{Y}$, then NN-MLM returns an optimal solution}

\begin{proof}
Recall that our loss function is $J(\mathbf{y}) = \sum_{k=1}^K \left(\|\mathbf{y} - \mathbf{t}_k\|^2 -  \hat{\delta}_k^2 \right)^2$. We can rewrite $J$ as:
\begin{align*}
    J(\mathbf{y}) = \sum_{l=1}^{2^L} N_l \left(\|\mathbf{y} - \mathbf{y}_l\|^2 -  \hat{\delta}_l^2 \right)^2
\end{align*}
where $\mathbf{y}_l$ is the $l$-th multilabel assignment, and $\hat{\delta}_l^2$ is the estimate of the distance between $\mathbf{y}$ and a reference point associated with $l$-th multilabel assignment $\mathbf{y}_l$. The value $N_l$ denotes the number of output reference points equal to $\mathbf{y}_l$. 
%Since we assume a perfectly balanced setting, $N_l=N$ for all $l$, and $J(\mathbf{y}) = \sum_{l=1}^{2^L} N \left(\|\mathbf{y} - \mathbf{y}_l\|^2 -  \hat{\delta}_l^2 \right)^2$.

%Thus, we have that
%\begin{align*}
%    &\argmin_{\mathbf{y} \in \mathbb{Y}} \sum_{l=1}^{2^L} N \left(\|\mathbf{y} - \mathbf{y}_l\|^2 -  \hat{\delta}_l^2 \right)^2 = \argmin_{\mathbf{y} \in \mathbb{Y}} \sum_{l=1}^{2^L} \left(\|\mathbf{y} - \mathbf{y}_l\|^2 -  \hat{\delta}_l^2 \right)^2
%\end{align*}
We are interested in the setting where $\hat{\delta}_l^2=\|\mathbf{y}' - \mathbf{y}_l\|^2$ for some $\mathbf{y}' \in \mathbb{Y}$, which gives
\begin{align*}
    &\argmin_{\mathbf{y} \in \mathbb{Y}} \sum_{l=1}^{2^L} N_l \left(\|\mathbf{y} - \mathbf{y}_l\|^2 - \|\mathbf{y}' - \mathbf{y}_l\|^2 \right)^2.
\end{align*}

A minimizer to the equation above is $\mathbf{y}=\mathbf{y}'$, which produces $J(\mathbf{y}')=0$. To conclude the proof, we only need to recover the value $\mathbf{y}'$ from $[\hat{\delta}_k]_k$. We achieve that by computing $\mathbf{y}' = \mathbf{t}_{k^\star}$ such that $\hat{\delta}_{k^\star}=0$, Notably, this corresponds to the prediction from the NN-MLM method.
\end{proof}

\section{Evaluation metrics} \label{app:metrics}

%Showing that a technique is better than some other model for a particular dataset(s) is often narrowed down by using just one evaluation metric and often with one predefined training and testing dataset division given in the literature. Thus it is unsurprising that supervised machine learning, especially single-label classification, has been so popular topic in machine learning. 

% Introduction to MLC evaluation with comparison to SLC
In the traditional single-label classification SLC, comparing different machine learning methods' classification performance is straightforward. In SLC, an input for the evaluation metrics from the machine learning model is just a set of labels with the same label cardinality as the test set. Many machine learning models have an intermediate layer that gives predicted scores for individual labels, which are then utilized to predict the relevant label by selecting the label corresponding to the largest score. Typically, these intermediate scores are not used with the SLC evaluation metrics. From the perspective of a classifier model with this kind of intermediate layer, MLC and SLC differ in the prediction phase only by how the intermediate values are handled. In MLC, we do not have a global fixed value for the number of relevant labels per instance. The MLC metrics can be divided into two main categories: $1)$ ranking-based and $2)$ bipartition-based. The bipartition-based metrics can be further divided into two subcategories: $2.1)$ example-based and $2.2)$ label-based.

%, the label ranking defined by the intermediate scores and thresholded ranking/scores bipartition are evaluated with separate metrics. These two main categories for the MLC evaluation metrics are $1)$ ranking-based and $2)$ bipartition-based. The bipartition-based metrics can be further divided into two subcategories: $2.1)$ example-based and $2.2)$ label-based.

% Introduce notations for example based metrics
For a test set of instances $\X_{ts} = \{\mathbf{x}_i^*\}_{i=1}^{N_{ts}}$ and a set of ground truth labels $\Y_{gt} = \{\mathbf{g}_i\}_{i=1}^{N_{ts}}$, where $\X_{ts} \subset \Rea^{M}$ and $\Y_{gt} \subset \{0,1\}^{L}$. For a classifier model $h$, we get a set of multi-label predictions $\Y_{pr} = \{\mathbf{y}_i^*\}_{i=1}^{N_{ts}}$, where $\mathbf{y}_i^* = h(\mathbf{x}_i^*) \in \{0,1\}^{L}$.

%Let $\Y = \{\vec{y}_i\}_{i=1}^N$ a set of label vectors for $\X$ so that $\vec{y}_i \in \{0,1\}^{L}$, where $\vec{y}_i(j) = 1$ associates that an instance $\vec{x}_i$ belongs to a class $j$.

% Introduce ranking-based metrics and their formulas
Hamming Loss is an example-based evaluation metric and it is one of the most popular MLC metrics. It is defined as
\begin{equation}
\label{eq:hammingloss}
\textsc{Hamming Loss} =  \frac{1}{N_{ts}L} \sum_{i = 1}^{N_{ts}}  \mathbf{y}_{i}^* \Delta \mathbf{g}_i
\end{equation}
%\begin{equation}
%\label{eq:hammingloss}
%hl =  \frac{1}{N_{ts}L} \sum_{i = 1}^{N_{ts}}  \sum_{j = 1}^{L} , \vec{y}_{i(j)}^* \oplus \vec{g}_{i(j)}
%\end{equation}
%where $\vec{g}_i \in \{0,1\}^{L}$ denotes the ground truth labels for a instance $i$. 
where $\Delta$ denotes the symmetric difference, which is defined as $\sum_{j = 1}^{L} \mathbf{a}_{(j)} \oplus \mathbf{b}_{(j)}$, for $\mathbf{a}, \mathbf{b} \in \{0,1\}^{L}$. In the MLC problems, accuracy is defined as
\begin{equation}
\label{eq:accuracy}
\textsc{Accuracy} =  \frac{1}{N_{ts}} \sum_{i = 1}^{N_{ts}} \frac{ \mathbf{y}_{i}^* \cdot \mathbf{g}_{i}}{\mathbf{y}_{i}^* \cdot \mathbf{y}_{i}^* + \mathbf{g}_{i} \cdot \mathbf{g}_{i} - \mathbf{y}_{i}^* \cdot \mathbf{g}_{i} }
\end{equation}
Micro recall and micro precision metrics are defined as
\begin{equation}
\label{eq:micro_recall}
\textsc{Micro Recall} = \frac{\sum_{j = 1}^{L} TP_j}{ \sum_{j = 1}^{L} TP_j + 
 \sum_{j = 1}^{L} FP_j}, 
\end{equation}
\begin{equation}
\label{eq:micro_precision}
\textsc{Micro Precision} = \frac{\sum_{j = 1}^{L} TP_j}{\sum_{j = 1}^{L} TP_j + \sum_{j = 1}^{L} FN_j},
\end{equation}
where, for a label $j$ and over $N_{ts}$ instances, $TP_j$ refers to \#True Positives, $FP_j$ to \#False Positives, and $FN_j$ to \#False Negatives. Corresponding macro metrics are defined as
\begin{equation}
\label{eq:macro_recall}
\textsc{Macro Recall} = \frac{1}{L} \sum_{j = 1}^{L} \frac{TP_j}{TP_j + FP_j}, 
\end{equation}
\begin{equation}
\label{eq:macro_precision}
\textsc{Macro Precision} = \frac{1}{L} \sum_{j = 1}^{L}  \frac{TP_j}{TP_j + FN_j},
\end{equation}
where $TP_j$, $FP_j$, and $FN_j$ are same as for the micro metrics (Equations \eqref{eq:micro_recall}-\eqref{eq:micro_precision}). Besides \textsc{Hamming Loss}, other popular bipartition-based MLC metrics are Micro F1 and Macro F1 \cite{pereira2018correlation, madjarov2012extensive} which are defined via the harmonic mean of the corresponding recall and precision as
\begin{equation}
\label{eq:micro_f1}
\textsc{Micro F1} = \textsc{Hmean} ~ (\textsc{Micro Precision},\textsc{Micro Recall}), 
\end{equation}
\begin{equation}
\label{eq:macro_f1}
\textsc{Macro F1} = \textsc{Hmean} ~ (\textsc{Macro Precision},\textsc{Macro Recall}), 
\end{equation}
where $\textsc{Hmean}(a,b) = \frac{2ab}{a+b}$ for $a,b \in \Rea_+$.

%where $TP_j = \sum_{i = 1}^{N} \mathbf{y}_{i(j)}^* \cdot \mathbf{g}_{i(j)}$ (\#True Positives for a label $j$), $FP_j = \sum_{i = 1}^{N} \mathbf{g}_{i(j)} \cdot \mathbf{g}_{i(j)} - \mathbf{y}_{i(j)}^* \cdot \mathbf{g}_{i(j)}$ (\#False Positives for a label $j$), and $FN_j = \sum_{i = 1}^{N} \mathbf{y}_{i(j)}^* \cdot \mathbf{y}_{i(j)}^* - \mathbf{y}_{i(j)}^* \cdot \mathbf{g}_{i(j)}$ (\#False Negatives for a label $j$).

%\begin{equation}
%\label{eq:accuracy}
%\textsc{Accuracy} =  \frac{1}{N_{ts}} \sum_{i = 1}^{N_{ts}}   \frac{\vert \mathbf{y}_{i}^* \cap \mathbf{g}_{i} \vert}{\vert \mathbf{y}_{i}^* \cup \mathbf{g}_{i} \vert}
%\end{equation}

% Define notations for ranking based metrics
For the given classifier $h$, lets denote $h_0$ as function that assigns a label score for each label. Then, for the classifier $h$, a predicted set of label scores $\mathbf{Z} = \{\mathbf{z}_i\}_{i=1}^{N_{test}}$, where $\mathbf{z}_i = h_0(\mathbf{x}_i) \in \Rea^{L}$ and $\mathbf{x}_i \in \X_{test}$.

% Introduce ranking-based metrics and their formulas
Ranking loss is a ranking-based MLC evaluation metric that represents overall quality of a classifier independently from the threshold selection. The ranking loss metric is given as
\begin{equation}
\label{eq:ranking_loss}
\textsc{Ranking Loss} =  \frac{1}{N_{ts}} \sum_{i = 1}^{N_{ts}} \frac{\vert \{ (j,k) ~\vert~ \mathbf{z}_{i(j)} < \mathbf{z}_{i(k)}, ~j \in g_i^+, ~ k \in g_i^0 \}\vert}{ \vert g_i^+ \vert \vert  g_i^0 \vert},
\end{equation}
where $g_i^+$ and $g_i^0$ are sets of the relevant and irrelevant ground truth labels, correspondingly. 
%where $g_i^+ = \{ j: \mathbf{g}_{i(j)} = 1 \}$ is a set of label indices referring to relevant ground truth labels for an instance $i$, and, correspondingly, $g_i^0 = \{ j: \mathbf{g}_{i(j)} = 0 \}$ is a set of label indices refererring to irrelevant ground ground truth labels for an instance $i$.
Another popular ranking-based metric, coverage, is defined as
\begin{equation}
\label{eq:coverage}
\textsc{Coverage} = \frac{1}{N_{ts}} \sum_{i = 1}^{N_{ts}} \vert \{j ~\vert ~ \min \limits_{k \in g_i^+} \mathbf{z}_{i(k)} \leq \mathbf{z}_{i(j)} \} \vert,
\end{equation}
where $g_i^+ = \{j~ \vert ~ j = \{ 1,...,L\},  g_{i(j)}=1\}$ denotes the active ground truth labels for an instance $i$. In short, this metric represents steps required to capture all actual labels from the ranking.

One error represents missclassification error for the highest ranked label. It is defined as
\begin{equation}
\label{eq:one_error}
\textsc{One error} = \frac{1}{N_{ts}} \sum_{i = 1}^{N_{ts}} 1 - \mathbf{g}_{i(j^*)},
\end{equation}
where $j^* = \argmax \limits_{j} \mathbf{z}_{i(j)}$. Note that \textsc{One error} possess the same information as the precision at $k=1$ ($P@1$) metric, since $\textsc{One error} = 1 - P@1$.

Average precision is defined as
\begin{equation}
\label{eq:average_precision}
\textsc{Average Precision} = \frac{1}{N_{ts}} \sum_{i = 1}^{N_{ts}} \frac{1}{\vert  g_i^+ \vert} \sum_{j \in g_i^+} \frac{\vert \{m ~\vert ~ \mathbf{z}_{i(j)} \leq \mathbf{z}_{i(m)}, m \in g_i^+ \} \vert}{\vert \{k ~\vert ~ \mathbf{z}_{i(j)} \leq \mathbf{z}_{i(k)} \} \vert}.
\end{equation}

This metric computes an average ratio of the relevant labels ranked above a label $j$ over \textsc{Coverage} plus one for the label $j$ (coverage above).

\section{Power parameter effects} \label{app:demo}

\begin{figure}[h]
%\captionsetup[subfigure]
  \centering
  \begin{subfigure}{0.45\textwidth}
    \includegraphics[width=\textwidth]{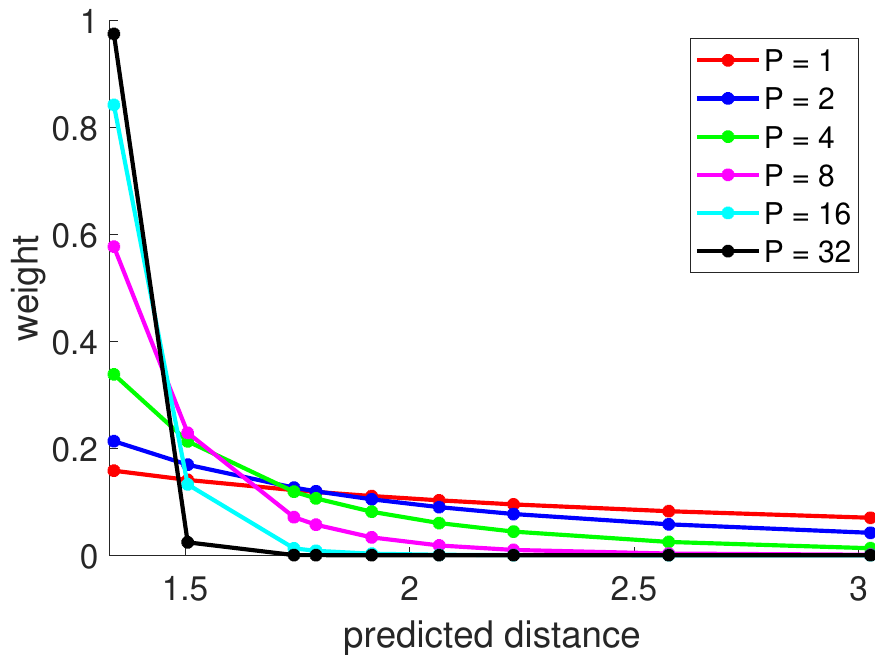}
    \caption{}
    \label{fig:weights}
  ~
  \end{subfigure}
  \begin{subfigure}{0.45\textwidth}
    \includegraphics[width=\textwidth]{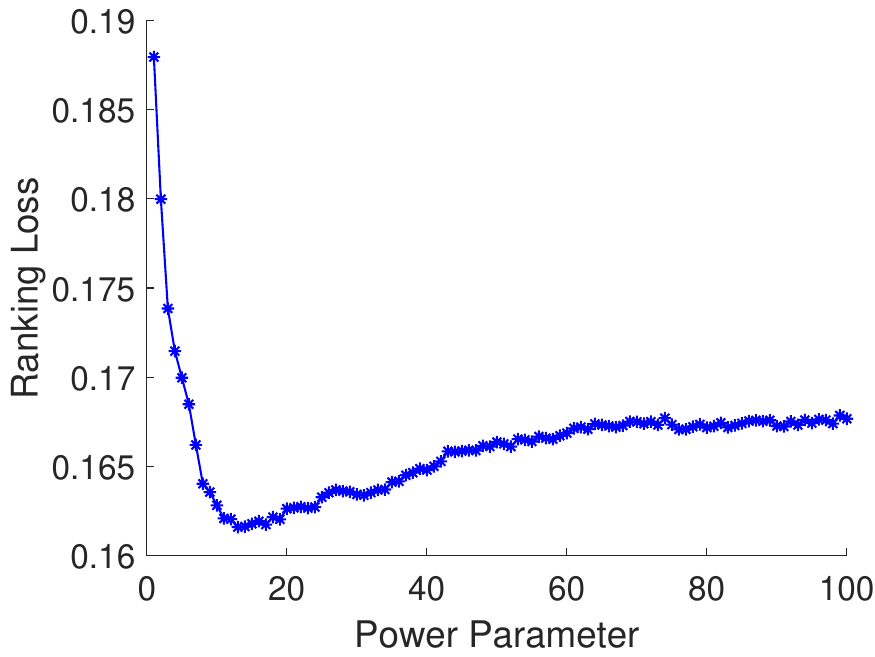}
        \caption{}
    \label{fig:yeastdemo}
  \end{subfigure}
  
	    \caption{(a) shows how weight values changes as a function MLM's predicted distances with different power parameter values. (b) shows how \textsc{Ranking Loss} changes as a function of power parameter $P$ for Yeast dataset.}
	\label{fig:illustration}
\end{figure}

% A) An illustration for sampled set of predicted distances from the actual model used for Yeast dataset. B) Yeast dataset demo: description how the ranking is affected for with different P values. 
Figure \ref{fig:illustration} shows how the power parameter affects to the weights and overall ranking performance. In Figure \ref{fig:weights}, weights for a sample of eight predicted distances are shown with $P = 1,2,4,8,16,32$. This set of predicted distances is sampled from a MLM model's prediction for the Yeast dataset. Note that since we are using the Euclidean distance as a distance measure and the binary representation of the labels, distance $\sqrt{m}$ corresponds to $m$ label difference. In the illustration, the smallest predicted distance is near to $\sqrt{2}$ which corresponds to two label difference. Note that this gives estimation how much the model is extrapolating. Increasing the power parameter value, gives relatively more weight to smaller predicted distances, meaning that the prediction is relying more on a predictions given by smaller set reference points. In Figure \ref{fig:yeastdemo}, the proposed method's \textsc{Ranking Loss} (see eq. \eqref{eq:ranking_loss}) is shown as a function of $P$ for a Yeast dataset's test set. For a smaller $P$ values, \textsc{Ranking Loss} decreases until it reaches the optimal value at $P = 13$. After this point, the \textsc{Ranking Loss} starts to slightly increase, which indicates that smaller predicted distances are being given too much weight in the convex combination of label vectors, causing deviations from the optimal ranking of the labels.

\newpage
\section{Results for ranking-based metrics}\label{app:ranking}

 \begin{table}[h]
 \caption{Results for \textsc{Ranking Loss}}\label{tab:rl}
 \normalsize
 \centering
 \bgroup
 \def\arraystretch{1.2}%
 \begin{adjustbox}{max width=1\columnwidth}
 \begin{tabular}{lrrrrrrrrrrrr}
 \hline
Dataset & Medical & Emotions & Enron & Scene & Yeast & Corel5k & Bibtex & Delicious & Tmc2007 & Mediamill\\ \hline
BR-SVM & 0.021 & 0.246 & 0.084 & \bf{0.060} & \bf{0.164} & 0.117 & 0.068 & 0.114 & 0.003 & 0.061 \\
CC-SVM & \bf{0.019} & 0.245 & 0.083 & 0.064 & 0.170 & 0.118 & \bf{0.067} & 0.117 & 0.003 & 0.062 \\
HOMER & 0.090 & 0.297 & 0.183 & 0.119 & 0.205 & 0.352 & 0.255 & 0.379 & 0.028 & 0.177 \\
RF-PCT & 0.024 & 0.151 & \bf{0.079} & 0.072 & 0.167 & 0.117 & 0.093 & \bf{0.106} & 0.006 & \bf{0.047} \\
ML-kNN & 0.045 & 0.283 & 0.093 & 0.093 & 0.172 & 0.130 & 0.217 & 0.129 & 0.031 & 0.055 \\
BR-MLM & 0.024 & 0.146 & 0.089 & 0.066 & 0.167 & 0.184 & 0.080 & 0.132 & \bf{0.000} & 0.061 \\
ML-MLM & 0.030 & \bf{0.142} & 0.081 & 0.065 & 0.166 & \bf{0.115} & 0.078 & 0.118 & \bf{0.000} & 0.051 \\
LLS-MLM & 0.029 & 0.155 & 0.111 & 0.068 & 0.174 & 0.194 & 0.089 & 0.148 & \bf{0.000} & 0.082 \\
\hline
\end{tabular}
 \end{adjustbox}
 \egroup
 \end{table}

 \begin{table}[!ht]
 \caption{Results for \textsc{Coverage}.}\label{tab:cov}
 \normalsize
 \centering
 \bgroup
 \def\arraystretch{1.2}%
 \begin{adjustbox}{max width=1\columnwidth}
 \begin{tabular}{lrrrrrrrrrrrr}
 \hline
Dataset & Medical & Emotions & Enron & Scene & Yeast & Corel5k & Bibtex & Delicious & Tmc2007 & Mediamill\\ \hline
BR-SVM & 1.610 & 2.307 & 12.530 & \bf{0.399} & 6.330 & 104.800 & \bf{20.926} & 530.126 & 1.311 & 20.481 \\
CC-SVM & \bf{1.471} & 2.317 & 12.437 & 0.417 & 6.439 & 105.428 & 21.078 & 537.388 & 1.302 & 20.333 \\
HOMER & 5.324 & 2.634 & 24.190 & 0.739 & 7.285 & 250.800 & 65.626 & 933.956 & 2.369 & 47.046 \\
RF-PCT & 1.619 & 1.827 & 12.074 & 0.461 & 6.179 & 107.412 & 25.854 & \bf{504.999} & 1.219 & \bf{16.926} \\
ML-kNN & 2.844 & 2.490 & 13.181 & 0.569 & 6.414 & 113.046 & 56.266 & 589.898 & 2.155 & 18.719 \\
BR-MLM & 1.592 & 1.792 & 14.143 & 0.439 & 6.339 & 157.048 & 25.883 & 664.122 & 1.208 & 22.412 \\
ML-MLM & 2.026 & \bf{1.743} & \bf{11.827} & 0.426 & \bf{6.022} & \bf{101.548} & 24.094 & 510.747 & \bf{1.207} & 17.956 \\
LLS-MLM & 2.008 & 1.847 & 16.057 & 0.450 & 6.610 & 166.308 & 29.010 & 715.828 & \bf{1.207} & 27.611 \\
\hline 
 \end{tabular}
 \end{adjustbox}
 \egroup
 \end{table}

 \begin{table}[h]
 \caption{Results for \textsc{One Error}.}\label{tab:oe}
 \normalsize
 \centering
 \bgroup
 \def\arraystretch{1.2}%
 \begin{adjustbox}{max width=1\columnwidth}
 \begin{tabular}{lrrrrrrrrrrrr}
 \hline
Dataset & Medical & Emotions & Enron & Scene & Yeast & Corel5k & Bibtex & Delicious & Tmc2007 & Mediamill\\ \hline
BR-SVM & 0.135 & 0.386 & 0.237 & \bf{0.180} & 0.236 & 0.660 & 0.346 & 0.354 & 0.029 & 0.188 \\
CC-SVM & \bf{0.123} & 0.376 & 0.238 & 0.204 & 0.268 & 0.674 & 0.342 & 0.367 & 0.026 & 0.193 \\
HOMER & 0.216 & 0.411 & 0.314 & 0.216 & 0.248 & 0.652 & 0.466 & 0.509 & 0.050 & 0.219 \\
RF-PCT & 0.174 & 0.262 & \bf{0.221} & 0.210 & 0.248 & 0.608 & 0.433 & 0.332 & 0.006 & 0.159 \\
ML-kNN & 0.279 & 0.406 & 0.280 & 0.242 & \bf{0.234} & 0.706 & 0.576 & 0.416 & 0.190 & 0.182 \\
BR-MLM & 0.153 & 0.267 & 0.230 & 0.194 & 0.236 & 0.600 & \bf{0.341} & \bf{0.309} & \bf{0.002} & \bf{0.113} \\
ML-MLM & 0.146 & \bf{0.257} & 0.295 & 0.195 & \bf{0.234} & 0.626 & 0.357 & 0.422 & \bf{0.002} & 0.121 \\
LLS-MLM & 0.158 & 0.262 & 0.275 & 0.191 & 0.236 & \bf{0.598} & 0.345 & 0.356 & \bf{0.002} & 0.129 \\
\hline 
 \end{tabular}
 \end{adjustbox}
 \egroup
 \end{table}
 
 \begin{table}[h]
 \caption{Results for \textsc{Average Precision}.}\label{tab:ap}
 \normalsize
 \centering
 \bgroup
 \def\arraystretch{1.2}%
 \begin{adjustbox}{max width=1\columnwidth}
 \begin{tabular}{lrrrrrrrrrrrr}
 \hline
Dataset & Medical & Emotions & Enron & Scene & Yeast & Corel5k & Bibtex & Delicious & Tmc2007 & Mediamill\\ \hline
BR-SVM & 0.896 & 0.721 & 0.693 & \bf{0.893} & \bf{0.768} & 0.303 & 0.597 & 0.351 & 0.978 & 0.686 \\
CC-SVM & \bf{0.901} & 0.724 & 0.695 & 0.881 & 0.755 & 0.293 & 0.599 & 0.343 & 0.981 & 0.672 \\
HOMER & 0.786 & 0.698 & 0.604 & 0.848 & 0.740 & 0.222 & 0.407 & 0.231 & 0.945 & 0.583 \\
RF-PCT & 0.868 & 0.812 & 0.698 & 0.874 & 0.757 & \bf{0.334} & 0.525 & 0.395 & 0.996 & 0.737 \\
ML-kNN & 0.784 & 0.694 & 0.635 & 0.851 & 0.758 & 0.266 & 0.349 & 0.326 & 0.844 & 0.703 \\
BR-MLM & 0.884 & 0.820 & \bf{0.709} & 0.883 & 0.766 & 0.327 & \bf{0.608} & \bf{0.407} & \bf{0.999} & \bf{0.745} \\
ML-MLM & 0.882 & \bf{0.827} & 0.685 & 0.883 & 0.767 & 0.321 & 0.587 & 0.345 & \bf{0.999} & 0.727 \\
LLS-MLM & 0.876 & 0.813 & 0.683 & 0.883 & 0.760 & 0.322 & 0.597 & 0.391 & \bf{0.999} & 0.724 \\
\hline 
 \end{tabular}
 \end{adjustbox}
 \egroup
 \end{table}

\newpage
\section{Results for bipartition-based metrics}\label{app:bipartition}

\begin{table}[ht]
 \caption{Results for \textsc{Accuracy}.}\label{tab:acc}
 \normalsize
 \centering
 \bgroup
 \def\arraystretch{1.2}%
 \begin{adjustbox}{max width=1\columnwidth}
 \begin{tabular}{lrrrrrrrrrrrr}
 \hline
Dataset & Medical & Emotions & Enron & Scene & Yeast & Corel5k & Bibtex & Delicious & Tmc2007 & Mediamill\\ \hline
BR-SVM & 0.206 & 0.361 & 0.446 & 0.689 & 0.520 & 0.030 & 0.348 & 0.136 & 0.891 & 0.403 \\
CC-SVM & 0.211 & 0.356 & 0.334 & 0.723 & 0.527 & 0.030 & 0.352 & 0.137 & 0.899 & 0.390 \\
HOMER & 0.713 & 0.471 & 0.478 & 0.717 & 0.559 & 0.179 & 0.330 & 0.207 & 0.888 & 0.413 \\
RF-PCT & 0.591 & 0.519 & 0.416 & 0.541 & 0.478 & 0.009 & 0.166 & 0.146 & 0.914 & 0.441 \\
ML-kNN & 0.528 & 0.319 & 0.319 & 0.629 & 0.492 & 0.014 & 0.129 & 0.102 & 0.574 & 0.421 \\
BR-MLM & 0.767 & 0.598 & 0.470 & 0.767 & 0.545 & 0.168 & 0.393 & 0.150 & 0.992 & 0.463 \\
ML-MLM & 0.762 & \bf{0.609} & 0.473 & 0.764 & \bf{0.568} & \bf{0.197} & \bf{0.411} & 0.223 & \bf{0.994} & 0.466 \\
LLS-MLM & 0.767 & 0.586 & 0.465 & \bf{0.770} & 0.543 & 0.173 & 0.400 & 0.149 & 0.993 & 0.465 \\
NN-MLM & \bf{0.775} & 0.586 & 0.455 & \bf{0.770} & 0.553 & 0.178 & 0.399 & 0.143 & 0.993 & \bf{0.467} \\
BPNN & 0.543 & 0.283 & 0.344 & 0.200 & 0.518 & 0.134 & 0.100 & 0.149 & 0.327 & 0.349 \\
BR-Ada & 0.731 & 0.458 & 0.384 & 0.609 & 0.500 & 0.015 & 0.283 & 0.087 & 0.481 & 0.391 \\
BR-RF & 0.445 & 0.494 & 0.443 & 0.556 & 0.495 & 0.050 & 0.192 & 0.188 & 0.993 & 0.448 \\
EBR-J48 & 0.655 & 0.519 & 0.479 & 0.648 & 0.511 & 0.139 & 0.343 & \bf{0.224} & 0.828 & 0.443 \\
ECC-J48 & 0.681 & 0.537 & \bf{0.480} & 0.659 & 0.519 & 0.129 & 0.337 & 0.188 & 0.820 & 0.429 \\
ML-kNN(d) & 0.457 & 0.357 & 0.356 & 0.651 & 0.479 & 0.050 & 0.154 & 0.137 & 0.587 & 0.418 \\
Rakel-SGDC & 0.646 & 0.349 & 0.380 & 0.600 & 0.507 & 0.103 & 0.318 & 0.134 & 0.612 & 0.402 \\
RakelSVC & 0.414 & 0.254 & 0.412 & 0.684 & 0.507 & 0.022 & 0.203 & 0.117 & 0.725 & 0.399 \\
\hline 
\end{tabular}
\end{adjustbox}
\egroup
\end{table}

\begin{table}[ht]
\caption{Results for \textsc{Hamming loss}.}\label{tab:hl}
\normalsize
\centering
\bgroup
\def\arraystretch{1.2}%
\begin{adjustbox}{max width=1\columnwidth}
\begin{tabular}{lrrrrrrrrrrrr}
\hline
Dataset & Medical & Emotions & Enron & Scene & Yeast & Corel5k & Bibtex & Delicious & Tmc2007 & Mediamill\\ \hline
BR-SVM & 0.077 & 0.257 & \bf{0.045} & 0.079 & \bf{0.190} & 0.017 & \bf{0.012} & \bf{0.018} & 0.013 & 0.032 \\
CC-SVM & 0.077 & 0.256 & 0.064 & 0.082 & 0.193 & 0.017 & \bf{0.012} & \bf{0.018} & 0.013 & 0.032 \\
HOMER & 0.012 & 0.361 & 0.051 & 0.082 & 0.207 & 0.012 & 0.014 & 0.022 & 0.015 & 0.038 \\
RF-PCT & 0.014 & 0.189 & 0.046 & 0.094 & 0.197 & \bf{0.009} & 0.013 & \bf{0.018} & 0.011 & \bf{0.029} \\
ML-kNN & 0.017 & 0.294 & 0.051 & 0.099 & 0.198 & \bf{0.009} & 0.014 & \bf{0.018} & 0.058 & 0.031 \\
BR-MLM & \bf{0.011} & 0.194 & 0.047 & 0.078 & 0.193 & 0.010 & 0.013 & \bf{0.018} & \bf{0.001} & \bf{0.029} \\
ML-MLM & 0.013 & \bf{0.186} & 0.054 & 0.078 & 0.195 & 0.014 & 0.018 & 0.025 & \bf{0.001} & 0.036 \\
LLS-MLM & \bf{0.011} & 0.201 & 0.049 & \bf{0.077} & 0.195 & 0.010 & 0.013 & \bf{0.018} & \bf{0.001} & \bf{0.029} \\
NN-MLM & \bf{0.011} & 0.202 & 0.050 & \bf{0.077} & 0.193 & 0.010 & 0.013 & \bf{0.018} & \bf{0.001} & \bf{0.029} \\
BPNN & 0.022 & 0.407 & 0.072 & 0.314 & 0.215 & 0.015 & 0.028 & 0.029 & 0.117 & 0.046 \\
BR-Ada & \bf{0.011} & 0.238 & 0.048 & 0.099 & 0.205 & \bf{0.009} & 0.013 & \bf{0.018} & 0.071 & 0.032 \\
BR-RF & 0.017 & 0.202 & 0.046 & 0.092 & 0.195 & 0.010 & 0.013 & \bf{0.018} & \bf{0.001} & \bf{0.029} \\
EBR-J48 & 0.015 & 0.216 & 0.051 & 0.099 & 0.217 & 0.015 & 0.017 & 0.024 & 0.023 & 0.036 \\
ECC-J48 & 0.014 & 0.236 & 0.056 & 0.109 & 0.233 & 0.023 & 0.018 & \bf{0.018} & 0.025 & 0.041 \\
ML-kNN(d) & 0.018 & 0.304 & 0.053 & 0.090 & 0.209 & 0.010 & 0.015 & 0.019 & 0.056 & 0.033 \\
Rakel-SGDC & 0.014 & 0.400 & 0.057 & 0.143 & 0.204 & 0.012 & 0.014 & \bf{0.018} & 0.053 & 0.032 \\
Rakel-SVC & 0.018 & 0.337 & 0.046 & 0.085 & 0.193 & \bf{0.009} & 0.013 & \bf{0.018} & 0.036 & 0.031 \\
\hline 
\end{tabular}
\end{adjustbox}
\egroup
\end{table}

\begin{table}[ht]
 \caption{Results for \textsc{Micro f1}.}\label{tab:microf1}
 \normalsize
 \centering
 \bgroup
 \def\arraystretch{1.2}%
 \begin{adjustbox}{max width=1\columnwidth}
 \begin{tabular}{lrrrrrrrrrrrr}
 \hline
Dataset & Medical & Emotions & Enron & Scene & Yeast & Corel5k & Bibtex & Delicious & Tmc2007 & Mediamill\\ \hline
BR-SVM & 0.343 & 0.509 & 0.564 & 0.761 & 0.652 & 0.059 & 0.457 & 0.234 & 0.932 & 0.533 \\
CC-SVM & 0.350 & 0.503 & 0.482 & 0.757 & 0.650 & 0.059 & \bf{0.462} & 0.236 & 0.936 & 0.509 \\
HOMER & 0.773 & 0.588 & 0.591 & 0.764 & 0.673 & 0.275 & 0.429 & 0.339 & 0.927 & 0.553 \\
RF-PCT & 0.693 & 0.672 & 0.537 & 0.669 & 0.617 & 0.018 & 0.230 & 0.248 & 0.945 & 0.563 \\
ML-kNN & 0.634 & 0.457 & 0.466 & 0.661 & 0.625 & 0.030 & 0.206 & 0.175 & 0.682 & 0.545 \\
BR-MLM & 0.788 & 0.705 & 0.580 & 0.775 & 0.663 & 0.256 & 0.437 & 0.275 & 0.994 & 0.586 \\
ML-MLM & 0.765 & \bf{0.715} & 0.571 & \bf{0.781} & \bf{0.678} & \bf{0.285} & 0.408 & 0.348 & \bf{0.996} & \bf{0.588} \\
LLS-MLM & 0.784 & 0.695 & 0.570 & 0.778 & 0.659 & 0.259 & 0.435 & 0.272 & 0.994 & 0.583 \\
NN-MLM & 0.794 & 0.693 & 0.554 & 0.778 & 0.663 & 0.261 & 0.432 & 0.260 & 0.995 & 0.585 \\
BPNN & 0.603 & 0.386 & 0.479 & 0.229 & 0.645 & 0.217 & 0.165 & 0.265 & 0.423 & 0.495 \\
BR-Ada & \bf{0.796} & 0.604 & 0.528 & 0.703 & 0.630 & 0.031 & 0.395 & 0.152 & 0.603 & 0.520 \\
BR-RF & 0.587 & 0.647 & 0.566 & 0.682 & 0.629 & 0.090 & 0.262 & 0.312 & 0.995 & 0.568 \\
EBR-J48 & 0.739 & 0.657 & \bf{0.601} & 0.722 & 0.642 & 0.224 & 0.452 & \bf{0.366} & 0.888 & 0.583 \\
ECC-J48 & 0.755 & 0.664 & 0.595 & 0.710 & 0.648 & 0.209 & 0.444 & 0.312 & 0.879 & 0.568 \\
ML-kNN(d) & 0.582 & 0.477 & 0.491 & 0.723 & 0.615 & 0.096 & 0.255 & 0.237 & 0.695 & 0.544 \\
Rakel-SGDC & 0.729 & 0.475 & 0.512 & 0.639 & 0.636 & 0.173 & 0.416 & 0.228 & 0.713 & 0.526 \\
Rakel-SVC & 0.564 & 0.319 & 0.534 & 0.742 & 0.638 & 0.045 & 0.267 & 0.202 & 0.806 & 0.519 \\
\hline 
 \end{tabular}
 \end{adjustbox}
 \egroup
 \end{table}

\begin{table}[ht]
 \caption{Results for \textsc{Macro f1}.}\label{tab:macrof1}
 \normalsize
 \centering
 \bgroup
 \def\arraystretch{1.2}%
 \begin{adjustbox}{max width=1\columnwidth}
 \begin{tabular}{lrrrrrrrrrrrr}
 \hline
Data set & Medical & Emotions & Enron & Scene & Yeast & Corel5k & Bibtex & Delicious & Tmc2007 & Mediamill\\ \hline
BR-SVM & 0.361 & 0.440 & 0.143 & 0.765 & 0.392 & 0.021 & 0.307 & 0.096 & 0.942 & 0.056 \\
CC-SVM & 0.371 & 0.420 & 0.153 & 0.762 & 0.390 & 0.021 & 0.316 & 0.100 & 0.947 & 0.052 \\
HOMER & 0.282 & 0.570 & 0.167 & 0.768 & \bf{0.447} & 0.036 & 0.266 & 0.103 & 0.924 & 0.073 \\
RF-PCT & 0.207 & 0.650 & 0.122 & 0.658 & 0.322 & 0.004 & 0.055 & 0.083 & 0.857 & 0.112 \\
ML-kNN & 0.192 & 0.385 & 0.087 & 0.692 & 0.336 & 0.010 & 0.065 & 0.051 & 0.493 & 0.113 \\
BR-MLM & 0.297 & 0.694 & 0.216 & 0.780 & 0.422 & 0.038 & 0.265 & 0.159 & 0.993 & 0.180 \\
ML-MLM & 0.315 & \bf{0.703} & 0.212 & \bf{0.789} & 0.406 & 0.038 & 0.299 & 0.160 & \bf{0.994} & 0.193 \\
LLS-MLM & 0.303 & 0.683 & \bf{0.218} & 0.783 & 0.422 & 0.040 & 0.266 & \bf{0.162} & 0.993 & 0.192 \\
NN-MLM & 0.307 & 0.681 & 0.213 & 0.783 & 0.423 & 0.042 & 0.259 & 0.160 & 0.993 & \bf{0.195} \\
BPNN & 0.209 & 0.183 & 0.060 & 0.115 & 0.385 & 0.017 & 0.047 & 0.030 & 0.345 & 0.040 \\
BR-Ada & \bf{0.383} & 0.591 & 0.152 & 0.707 & 0.346 & 0.007 & 0.247 & 0.054 & 0.451 & 0.052 \\
BR-RF & 0.129 & 0.628 & 0.180 & 0.676 & 0.347 & 0.007 & 0.080 & 0.158 & 0.993 & 0.111 \\
EBR-J48 & 0.238 & 0.644 & 0.191 & 0.731 & 0.417 & 0.040 & 0.347 & 0.128 & 0.833 & 0.177 \\
ECC-J48 & 0.258 & 0.657 & 0.204 & 0.722 & 0.436 & \bf{0.045} & \bf{0.349} & 0.158 & 0.828 & 0.187 \\
MLkNN(d) & 0.166 & 0.426 & 0.100 & 0.724 & 0.397 & 0.027 & 0.122 & 0.078 & 0.592 & 0.150 \\
Rakel-SGDC & 0.273 & 0.424 & 0.199 & 0.649 & 0.363 & 0.042 & 0.281 & 0.088 & 0.604 & 0.036 \\
Rakel-SVC & 0.095 & 0.107 & 0.115 & 0.745 & 0.360 & 0.006 & 0.067 & 0.051 & 0.694 & 0.035 \\
\hline
\end{tabular}
\end{adjustbox}
\egroup
\end{table}

%An appendix contains supplementary information that is not an essential part of the text itself but which may be helpful in providing a more comprehensive understanding of the research problem or it is information that is too cumbersome to be included in the body of the paper.




\end{appendices}

\end{document}